\documentclass[oneside]{scrreprt}

\usepackage[top=2in,
            bottom=2in,
            left=1.5in,
            right=1.5in]{geometry}

\usepackage[phd]{tugrazthesis}
\usepackage{fancyhdr}
\usepackage[hidelinks]{hyperref}
\usepackage{pdfpages}
\usepackage{caption}
\usepackage{subcaption}
\usepackage{amsmath} 
\usepackage{enumitem}

\usepackage{tabularx}
\usepackage{booktabs}

\usepackage[backend=biber, style=ieee, maxbibnames=99]{biblatex}
\usepackage{cleveref}
\usepackage{epigraph}
\usepackage{csquotes}

\newlist{paperlist}{enumerate}{1}

\setlist[paperlist]{
    label=\textsc{Paper \arabic* --}, 
    ref=\arabic*, 
    align=left, 
    labelindent=0pt, 
    leftmargin=*
}

\crefname{paperlisti}{\textsc{paper}}{\textsc{papers}}
\Crefname{paperlisti}{\textsc{Paper}}{\textsc{Papers}}

\newlist{rqlist}{enumerate}{1}

\setlist[rqlist]{
    label=\textsc{RQ\arabic* --}, 
    ref=\arabic*, 
    align=left, 
    labelindent=0pt, 
    leftmargin=*
}

\crefformat{rqlisti}{#2\textsc{RQ}#1#3}
\Crefformat{rqlisti}{#2\textsc{RQ}#1#3}

\usepackage{xurl}
\usepackage[export]{adjustbox}

\usepackage{chngcntr}
\counterwithout{figure}{chapter}

\newcommand{\spara}[1]{\smallskip\noindent\textbf{#1}}

\addtokomafont{pagenumber}{\small\sffamily}
\begin{document}


\pagestyle{fancy}

\fancyhf{}

\fancyhead[C]{
    \sffamily\small
    \nouppercase{\thesection\quad\rightmark}
    \vspace{5pt}
    \par\hrule
}

\renewcommand{\headrulewidth}{0pt}

\renewcommand{\chaptermark}[1]{}
\renewcommand{\sectionmark}[1]{\markright{#1}}

\fancyfoot[C]{\sffamily\small\thepage}


\setlength{\parskip}{1em}

\RedeclareSectionCommand[beforeskip=1ex, afterskip=1ex]{chapter}
\RedeclareSectionCommand[beforeskip=1ex, afterskip=1ex]{section}
\RedeclareSectionCommand[beforeskip=0.8ex, afterskip=0.8ex]{subsection}
\RedeclareSectionCommand[beforeskip=0.8ex]{paragraph}

\newtheorem{definition}{Definition}


\thesisauthor[]{Antonio Ferrara}

\thesistitle[]{Statistical and Structural Approaches to \\ Algorithmic Fairness}


\thesisdate[ ]{19\textsuperscript{th} June 2026}

\supervisortitle{\germanenglish{Betreuerin/Betreuer}{Supervisor}}
\supervisor{Univ.-Prof. Dr. Fariba Karimi \\ Institute of Human-Centred Computing	}

\academicdegree{Doctor of Technical Sciences}


\printthesistitle

\printaffidavit

\chapter*{Abstract}

\begingroup
\markright{}
\renewcommand{\thesection}{}

Modern machine learning systems have outgrown their origins as isolated predictive constructs, evolving into complex socio-technical architectures that actively mediate human opportunity. As algorithms increasingly determine access to economic and social opportunities, it has become widely recognized that these systems are deeply embedded with the structural inequalities and prejudices of their environments. The field of algorithmic fairness emerged in response to the growing recognition that models optimized for predictive accuracy can systematically disadvantage marginalized groups. Early mitigation strategies, however, rested on fragile simplifications that limited their effectiveness in complex socio-technical environments. This thesis identifies and addresses two fundamental limitations of contemporary fairness paradigms: the reliance on deterministic point estimates for auditing and the treatment of individuals as isolated entities devoid of structural context.

First, the diagnosis of algorithmic unfairness has traditionally depended on scalar metrics that fail to capture the nuances of real-world deployment. This deterministic approach ignores the high statistical variance inherent in small, intersectional groups, often leading to false alarms or missed detections of bias. Furthermore, standard auditing struggles with the opacity of black-box models, frequently conflating unjustifiable bias with the influence of legitimate features. Moreover, traditional metrics also overwhelmingly focus on disparate outcomes, neglecting the procedural logic of the model and failing to detect when groups are subjected to distinct, discriminatory decision-making processes. To resolve these diagnostic failures, this thesis proposes a transition toward statistical hypothesis testing, ensuring that fairness assessments are statistically robust, causally valid, and capable of inspecting the underlying explanations of model decisions.

Second, this thesis challenges the prevailing focus on isolated predictions by examining algorithmic fairness through a structural lens. In networked  and hierarchical systems, fairness is an emergent property of the interactions, connectivity, and comparative processes that govern the domain. Network topologies are not neutral; when optimization objectives interact with these structures, they can actively concentrate visibility and marginalize already peripheral communities. Similarly, hierarchical structures and  rankings frequently crystallize human prejudices and biases into systemic disadvantages. By shifting the focus to these structural dependencies, the thesis demonstrates that achieving fairness requires deliberately reshaping how opportunities flow through networks and how merit is inferred and aggregated.

\clearpage
\thispagestyle{plain}

Ultimately, this thesis advances comprehensive frameworks that marry statistical reliability with structural awareness. By replacing brittle auditing metrics with statistical hypothesis testing, introducing concepts that actively mitigate bias in structural systems including physical routing, social networks, and rankings, and complementing these with operational safeguards such as abstention under uncertainty and lifecycle-wide bias governance, this thesis provides a robust foundation for the deployment of trustworthy artificial intelligence.

\endgroup
\clearpage

\chapter*{Acknowledgments}
I am deeply grateful to Prof. Fariba Karimi for supervising me throughout my entire Ph.D., providing a great combination of academic freedom and kind guidance. Her insightful advice and wise  feedback have been invaluable in shaping this work.

Furthermore, I extend my deepest gratitude to Prof. Claudia Wagner for her mentoring and support during the first years of my doctorate, and to Prof. Francesco Bonchi for his guidance and inspiration during the final years of my journey. 

I also want to thank all my  colleagues and friends from TU Graz, GESIS, NoBias, and CENTAI, for their wonderful friendship and intellectual stimulation.

A special thank you to my girlfriend, Veronica, for her endless patience while I wrote this thesis and for surviving stressful deadlines together, even a NeurIPS rebuttal during our vacation in  the Maldives! Grazie mille amorino.

Finally, my deepest thanks go to my family, for their unconditional support, for  believing in me, and for always giving me the freedom to follow the path I chose.

\cleardoublepage

\tableofcontents
\thispagestyle{plain}
\clearpage
\pagestyle{fancy}


\pagenumbering{arabic}

\chapter{Introduction}
\label{chap:introduction}

\begin{displayquote}
\textit{``Technology is neither good nor bad; nor is it neutral.''} \\
\hspace*{\fill} -- \textup{Melvin Kranzberg} \cite{kranzberg1986technology}
\end{displayquote}

\section{The Foundations and Challenges of Algorithmic Fairness}

The integration of Artificial Intelligence into the foundational infrastructure of modern society represents a technological shift of unprecedented magnitude. We live in an era where the most consequential decisions of our lives, how we find employment, how we access credit, how we consume information, and even how we navigate the physical world, are mediated by artificial intelligence.  Every day, billions of human needs are translated into queries, and billions of algorithmic responses shape the trajectory of those desires.

For the first few decades of the digital age, the primary challenge for these systems was accuracy: minimizing error rates on independent, identically distributed data points. Success was defined by a single number, e.g., accuracy, precision, or AUC, calculated on an often static test set. The optimization landscape was viewed as a technical frontier, devoid of moral weight. The underlying assumption was one of neutrality: if an algorithm is trained on data to maximize a mathematical objective function, it acts as an objective arbiter of quality.

This thesis proceeds from the fundamental premise that the era of assumed neutrality has irrevocably ended. As algorithmic systems have transitioned from passive tools to active gatekeepers of social and economic opportunity, they have begun to function less like technical instruments and more like institutions. They do not merely describe the world; they shape it. Today, algorithms determine who accesses credit, who is interviewed for employment, which neighborhoods receive infrastructure investment, and how information propagates through the public sphere.

A critical realization in the field of socio-technical systems is that an algorithm can be highly accurate on average while systematically discriminating against specific subpopulations. A credit scoring model might maximize overall profit while denying loans to creditworthy minority applicants due to historical redlining encoded in the training data. An online marketplace might boost overall sales by highlighting popular brands while burying highly rated products from small, minority-owned businesses. A route recommendation system might optimize for global traffic flow while economically strangling local businesses in specific neighborhoods by diverting all foot traffic away from them.   

The field of Algorithmic Fairness emerged to address these disparities, initially by defining statistical constraints such as Statistical Parity~\cite{dwork2012fairness} (requiring equal positive prediction rates across groups) or Equalized Odds~\cite{hardt2016equality} (requiring equal true positive and false positive error rates) to enforce equitable outcomes across groups defined by sensitive attributes like race, gender, or age. While foundational, this early era of fairness research relied on several critical simplifications that now limit the efficacy of auditing and intervention in complex, real-world scenarios. Specifically, this thesis aims to address two major limitations in the current literature: the overreliance on point estimates for fairness assessment, and the failure to account for the interdependence between individuals, particularly the structural relationships embedded within networks and hierarchies.

\subsection*{From Point Estimates to Fairness Testing}
\label{sec:intro_auditing_framework}

One of the first fundamental challenges in algorithmic fairness is the \textit{diagnosis} of the presence of inequalities. Before an algorithmic system can be corrected, its unfairness must be reliably measured. However, the operationalization of algorithmic fairness has traditionally relied on fragile simplifications, including the calculation of disparity as point estimates. In standard practice, a practitioner selects a metric, such as Statistical Parity \cite{dwork2012fairness} or Equal Opportunity~\cite{hardt2016equality}, computes it as a scalar value over a test set, and compares the result to a fixed threshold (e.g., the ``four-fifths rule''~\cite{greenberg1979analysis}).

This thesis argues that this deterministic, scalar approach is insufficient for the complexity of modern socio-technical systems. The reliance on simple point estimates fails to account for several critical dimensions of real-world deployment:
\begin{enumerate}
    \item \textbf{Reliability on Small Groups:} It ignores the high variance of estimators in finite, intersectional samples, leading to false alarms in small groups and missed detections in large ones.
    \item \textbf{Direct and Indirect Influences:} It cannot distinguish between unjustifiable bias and disparities explained by legitimate task-relevant features, especially in opaque ``black-box'' systems where the internal logic is inaccessible.
    \item \textbf{Process Transparency:} It measures only the final outcome distributions, failing to detect when a model arrives at equitable outcomes through discriminatory reasoning or proxy variables.
\end{enumerate}

In this thesis, we address these limitations by introducing  comprehensive auditing frameworks that transition from comparing point estimates to conducting statistical hypothesis tests. 

\spara{Reliability on Small Groups:} The first failure mode of standard auditing is the illusion of precision. As we disaggregate data to audit intersectional subgroups, defined by combinations of protected attributes such as race, gender, and age, sample sizes naturally shrink. As the granularity of the audit increases, the variance of fairness estimators explodes. 
In  \Cref{Pap:saft} (\textit{Size-adaptive Hypothesis Testing for Fairness}), we demonstrate that standard metrics frequently flag violations in intersectional groups that are essentially noise, while simultaneously missing significant disparities in larger groups due to rigid thresholds. We resolve this by reframing fairness auditing as a hypothesis testing problem. We introduce a size-adaptive framework that employs Wald tests for large samples and Bayesian inference for small samples. This approach allows us to define the \textit{resolution limit} of fairness: the minimal detectable disparity for a given subgroup size, effectively creating a ``no-power zone'' where data is insufficient to reject the null hypothesis of fairness.

\spara{Disentangling Direct and Indirect Influence:}
Even when sample sizes are sufficient, standard disparity metrics can be misleading because they often only measure marginal dependence rather than considering also conditional dependence. In many high-stakes ranking applications, such as hiring or credit scoring, the auditor typically has access to the input features, the protected attributes, and the output ranking, but not the internal scoring function of the model. A machine learning system might exhibit a correlation with a protected attribute (e.g., gender) simply because that attribute is correlated with a legitimate, task-relevant feature (e.g., education level). Standard metrics like Statistical Parity would flag this as unfair, potentially forcing a correction that reduces utility. Conversely, a model might be ostensibly fair on average but harbor ``residual'' bias where, for two individuals with identical task-relevant features, the protected group member is consistently ranked lower. In \Cref{Pap:condor} (\textit{Auditing for Demographic Bias in Opaque Rankings}), we propose a method to detect this residual dependence under strict black-box assumptions. We move beyond point estimates of representation to test the conditional independence hypothesis. We introduce a statistical framework called \textsc{Condor} that residualizes the ranking and protected attributes with respect to the observables using kernels in a Reproducing Kernel Hilbert Space \cite{scholkopf2002learning} (RKHS), and then quantifies the remaining association using distance correlation. This method allows auditors to distinguish between disparities rooted in direct or indirect influence of protected attributes.

\spara{Process Transparency:}
The final limitation of standard point estimates is their highly prevalent focus on \textit{outcomes} (Disparate Impact) rather than on the underlying \textit{process} (Disparate Treatment). Historically, doctrines such as ``fairness through unawareness'' attempted to ensure procedural fairness through strict blindness, i.e., by simply removing the protected attribute from the data. However, this naive approach fails to achieve true Equal Treatment (the formal absence of Disparate Treatment) because it ignores how sensitive attributes are indirectly encoded via proxies. Equal Treatment demands that a model's internal logic does not depend on group membership, either directly or indirectly. Crucially, a model can satisfy outcome-based metrics like Statistical Parity while still violating Equal Treatment. For example, a model might positively discriminate in favor of a subgroup based on a certain feature, while negatively discriminating against them on another feature, causing the biases to cancel out in the final prediction distribution. While the aggregated outcome appears fair, the underlying process is not; individuals are still being subjected to distinct, biased decision logic based on their demographic group. In \Cref{Pap:et} (\textit{Beyond Demographic Parity: Redefining Equal Treatment}), we introduce the metric of Explanation Disparity to diagnose this failure mode. Instead of comparing the distributions of predictions, we compare the distributions of \textit{explanations}, specifically, the Shapley value contributions of the features. We propose an Equal Treatment Inspector, a meta-classifier trained to predict the protected attribute solely from the explanations of the model's decisions. If this inspector can predict the protected attribute with accuracy better than random chance, it proves that the model is treating groups differently, even if the final outcome rates are identical.

Together, these three methodologies constitute a unified diagnostic layer for algorithmic fairness that far exceeds the capabilities of traditional point-estimate audits. By ensuring statistical validity against noise (\Cref{Pap:saft}), causal validity against confounding (\Cref{Pap:condor}), and procedural validity against proxy discrimination (\Cref{Pap:et}), we establish a robust foundation for the auditing of fairness strategies discussed in the subsequent chapters of this thesis.

\subsection*{From Isolated Individuals to Structured Systems}
\label{sec:intro_structural}

In the previous section, we explained problems and  challenges of auditing for fairness. A further limitation of traditional fairness lies in its treatment of individuals as isolated data points. Indeed, many contemporary algorithmic systems do not operate on independent individuals. Instead, they function over structured domains, where entities are connected, compared, ordered, or embedded in networks of mutual influence. In such environments, fairness is not solely a property of isolated predictions, but of the entire system of interactions through which outcomes are produced. 

In this thesis, we adopt a broad interpretation of the term \textbf{Structural}, referring to any setting in which relationships among entities play a central role in shaping outcomes. Such relationships may arise through graph connectivity, network diffusion processes, ranking positions, or pairwise comparisons. 
We argue that fairness cannot be fully understood by examining individuals and their characteristics in isolation; rather, it requires confronting the structural challenges, such as biased connectivity patterns, feedback loops, and distorted comparative processes, that govern these systems. 
Accordingly, this thesis extends the scope of algorithmic fairness beyond 
the analysis of independent predictions to encompass relational structures, identifying critical sources of failure in two fundamental forms of organization: \textbf{Network Structures}, which regulate how opportunities flow and become visible, and \textbf{Hierarchical Structures}, which determine how merit is inferred, ordered, and aggregated.

\subsubsection*{Unfairness in Network Structures}

In networked systems, algorithms do not merely predict individual outcomes, they actively shape the distribution of visibility, access, and opportunity across the underlying structure. Crucially, network topology is not neutral: it reflects historical constraints, spatial frictions, and social divisions that influence how advantages and disadvantages propagate. When optimization objectives focus narrowly on local efficiency or predictive accuracy, they interact with these structural properties to generate effects that progressively concentrate exposure and marginalize already peripheral nodes. As a result, inequality in networked systems is not simply the product of biased data or flawed predictions, but an emergent property of the interplay between topology and optimization, where visibility, connectivity, and opportunity become unevenly distributed. This structural perspective provides a unifying lens through which several seemingly distinct challenges can be understood.

In physical transportation networks, this dynamic appears as a tension between individual efficiency and systemic equity. Classical routing systems optimize for minimal distance or minimal travel time, yet this objective induces  winner-take-all patterns in which traffic, and therefore economic visibility, is funneled through a narrow set of “optimal” corridors, leaving nearby alternatives systematically deprived of exposure. The root cause is a structural bottleneck of optimality: in weighted graphs, the shortest path is frequently unique, and when an algorithm is constrained to always select it, fairness becomes structurally infeasible rather than merely overlooked. In \Cref{Pap:mmfp} (\textit{Beyond Shortest Paths: Node Fairness in Route Recommendation}) we overcome these limitations by relaxing strict optimality and expanding the feasible solution space through a novel definition of paths. Specifically, we generate routes that maintain a user's progress toward their destination without strictly minimizing distance. In this way, the exposure of business locations (the network nodes) to the customers traversing these paths can be redistributed toward the worst-off nodes while sufficiently preserving the overall user experience. In this sense, routing algorithms shift from passive optimizers of efficiency to active mediators of opportunities.

A closely related process governs social and recommendation networks, where the challenge is rooted in the dynamic  evolution of the network. Social graphs are shaped by homophily~\cite{mcpherson2001birds} and preferential attachment~\cite{barabasi1999emergence}, structural forces that naturally produce clustering and inequality in connectivity. When link recommendation systems rely on existing topology, optimizing for structural proximity naturally reproduces these patterns. As we demonstrate in \Cref{Pap:link} (\textit{Link Recommendations: Their Impact on Network Structure and Minorities}), this creates a severe feedback loop in which popular nodes become increasingly central while minority or peripheral groups remain isolated. Over time, the network itself becomes a barrier to opportunity, restricting access to information and social centrality. Crucially, we show that the objectives of purely topological similarity and structural fairness are often fundamentally misaligned: algorithms that prioritize within-cluster connections inadvertently intensify segregation. Addressing this requires more than re-ranking recommendations; it involves structural interventions that deliberately bridge disconnected regions of the graph, preventing persistent marginalization and counteracting network segregation effects.

These structural inequalities are further obscured by a granularity gap that limits our ability to observe them directly. In many geographical and social contexts, socioeconomic indicators are measured at coarse spatial or temporal resolution, masking fine-grained disparities that emerge from the actual flow of people, resources, and attention across networks. The challenge is therefore one of structural inference: how to extract localized inequality from weak or aggregated signals. Interaction graphs, such as user–business mobility networks, implicitly encode such information because movement patterns often exhibit homophily, linking individuals and places of similar socioeconomic status. In \Cref{Pap:super} (\textit{Super-Resolution of Urban Socioeconomic Indicators via Graph-Based Recommender Systems}), we study how to use Graph Neural Networks to propagate  latent signals across the network, effectively refining coarse observations and revealing otherwise invisible pockets of disadvantage. 

Taken together, these phenomena highlight a unifying principle: inequality in networked systems is not solely a consequence of biased data or imperfect predictions, but an emergent property of the interaction between topology and optimization. Achieving fairness therefore requires moving beyond local objective optimization toward algorithms that explicitly reason about, and when necessary reshape, the structure of connectivity and visibility that governs access to opportunity.

\subsubsection*{Unfairness in Hierarchical Structures}
Just as networks are structures of connection, rankings are structures of hierarchy. A major challenge in algorithmic fairness is that these hierarchies are rarely ``given''; they are often derived from noisy, biased preference  relationships. If the process of comparison itself is flawed, the resulting ranking will crystallize that bias into a systemic disadvantage.

The first point of failure lies in the input to these systems. Rankings are frequently aggregated from human judgments, which are rarely neutral. Evaluators often harbor implicit group biases, such as in-group favoritism or out-group prejudice, that distort their perception of an item's quality. Standard ranking models often fail to account for this, effectively assuming evaluators are objective sensors of latent quality. This conflation of prejudice with merit propagates discriminatory inputs directly into the final ranking, making the algorithmic output a reflection of human bias rather than true utility. In \Cref{Pap:barp} (\textit{Bias-Aware Ranking from Pairwise Comparisons}), we tackle this problem by developing a parametric method to account for evaluators' biases in rankings obtained from pairwise comparisons.

Even if evaluators were neutral, fairness is compromised by the process of candidate selection. In many systems, not all pairs are compared; sampling strategies determine \textit{who} gets the opportunity to be evaluated. Biases in sampling, where privileged individuals are selected for comparison more often due to popularity or representation bias, create a distorted comparison graph. Standard ranking recovery algorithms can interpret this lack of comparisons as a lack of merit, systematically under-ranking minority groups simply because they were denied the ``opportunity to compete''. We study this phenomenon in \Cref{Pap:georg} (\textit{Fairness-Aware Ranking Recovery from Pairwise Comparisons}).

Finally, the aggregation mechanism itself can become a vehicle for unfairness. When combining multiple preference lists into a consensus, algorithms often follow transition probabilities derived from the input data. If the original rankings contain position bias or consistently favor certain groups, the aggregation process will naturally converge to a consensus that marginalizes the other groups. The challenge is to intervene in these aggregation algorithms to ensure that the final consensus reflects a fair representation of all groups without faithfully reproducing the biases of the inputs. We develop such techniques in \Cref{Pap:fairMC} (\textit{FairMC Fair-Markov Chain Rank Aggregation Methods}).

By shifting the focus from isolated individuals to structural challenges, this thesis argues that fairness cannot be achieved merely by constraining final outcomes. It requires diagnosing and repairing the inputs (biased judgments), the processes (skewed sampling), and the mechanisms (aggregation and optimization) that constitute the system itself.

\section{Research Questions}\label{sec:RQ}

This thesis is grounded in the observation that algorithmic systems have transitioned from passive technical tools to active institutional gatekeepers that distribute social and economic opportunities, arguing that the traditional assumption of algorithmic neutrality has ended, as these systems now shape the world rather than merely describing it.

However, the current methods for ensuring fairness in these systems are failing due to several limitations. First, standard auditing relies on fragile ``point estimates'', scalar metrics that ignore statistical uncertainty and fail to detect bias in complex, black-box models. Furthermore, current approaches often treat individuals as isolated data points, ignoring the ``structural'' reality where opportunities are shaped by networks, connectivity, and hierarchies. Consequently, this research proposes a shift toward statistical validity (via hypothesis testing) and structural awareness (via network and ranking analysis). In more detail, in this  thesis we want to address the following research questions:

\begin{rqlist}[series=myrq]
    \item \label{rq:test} How can the diagnosis of algorithmic unfairness transition from point estimates to statistical hypothesis testing?
    \item \label{rq:networks} How do structural dependencies in networks create  inequalities, and how can they be mitigated?
    \item \label{rq:rank} How is unfairness embedded in hierarchical structures, and how can it be addressed?
    \item \label{rq:safety} How can systems be designed to ensure reliability and safety in the face of biased inputs and opaque logic?
\end{rqlist}

The research questions outlined above are addressed through the papers  presented in the following \Cref{sec:pub}.

\Cref{Pap:saft,Pap:condor,Pap:et} of this thesis address the first research question by proposing a comprehensive shift from deterministic point estimates to robust statistical hypothesis testing frameworks. To handle the high statistical variance inherent in small intersectional subgroups, \Cref{Pap:saft} introduces a unified hypothesis-testing approach utilizing Bayesian inference to provide reliable credible intervals when data is scarce. To tackle the influence of sensitive attributes, \Cref{Pap:condor} presents a model-agnostic auditing method that tests for conditional independence by measuring residual demographic dependence, effectively distinguishing between task-relevant features and unjustifiable bias. Lastly, \Cref{Pap:et} moves beyond disparate outcomes to inspect procedural logic. By comparing the distributions of feature attributions and training a meta-classifier to detect disparate treatment, the framework ensures models do not apply distinct, discriminatory reasoning across demographic groups, even when final acceptance rates appear equitable. 

To answer the second research question, the thesis demonstrates how inequalities naturally emerge when algorithms focus exclusively on narrow goals like efficiency or accuracy, often interacting with the underlying network to hide or marginalize vulnerable individuals or communities. In connected systems like navigation networks, \Cref{Pap:mmfp} of this thesis shows how to spread out exposure more fairly by recommending diverse, near-optimal routes instead of rigidly forcing everyone down a single shortest path. \Cref{Pap:link} contributes to \Cref{rq:networks} by inspecting how link recommendation algorithms, applied over time, can reinforce inequalities and segregation of communities, for example, by reducing the visibility of already marginalized minority groups. Instead, in \Cref{Pap:super}, we propose a GNN framework that learns business representations from a bipartite user-business interaction graph. The representations are then used to better infer socioeconomic indicators, and thereby reveal underlying socioeconomic inequalities.

To answer \Cref{rq:rank}, this thesis explores how unfairness permeates the entire pipeline of hierarchical structures, rankings, and pairwise comparisons. First, to address the distortion of human prejudice in the input data, \Cref{Pap:barp} proposes a probabilistic framework that explicitly estimates and mathematically corrects individual evaluator biases, allowing for the recovery of an unbiased latent ranking. Furthermore, \Cref{Pap:georg} investigates the ranking recovery phase, evaluating how skewed sampling strategies interact with standard and fairness-aware algorithms, demonstrating the need to prevent structural invisibility for underrepresented candidates. Finally, tackling the aggregation of multiple rankings into a consensus, \Cref{Pap:fairMC} introduces an in-processing Markov Chain method that rescales edge weights to ensure balanced visibility across demographic groups directly within the aggregation mechanism.

\Cref{Pap:baltor} bridges \Cref{rq:rank} and \Cref{rq:safety} by introducing a model-agnostic safety mechanism that enables ranking systems to selectively abstain from predictions when the underlying uncertainty is high. This is operationalized by estimating conditional risk and rejecting uncertain pairwise orderings without requiring the underlying model to be retrained.


Beyond direct algorithmic interventions, \Cref{Pap:policy} answers \Cref{rq:safety} by synthesizing technical findings into actionable governance, proposing a holistic bias management architecture. The proposed governance framework champions continuous monitoring for temporal distribution shifts, the integration of causal reasoning to explicitly model discriminatory mechanisms, and a cautious approach to explainable artificial intelligence to manage the vulnerabilities of opaque systems across the entire lifecycle.

\section{List of Publications}\label{sec:pub}
\subsection*{Publications in International Journals and Conferences (Published and Forthcoming)}
\subsubsection{As First Author}
\begin{paperlist}[series=mypapers]

    \item \label{Pap:saft}  \textit{Size-adaptive Hypothesis Testing for Fairness}~\cite{ferrara2025size}   \newline \textbf{A. Ferrara}, F. Cozzi, A. Perotti, A. Panisson, and F. Bonchi
    \newline Advances in Neural Information Processing Systems 38 (NeurIPS~2025)  

    \item \label{Pap:condor} \textit{Auditing for Demographic Bias in Opaque Rankings}~\cite{ferra2025audit}
    \newline \textbf{A. Ferrara}, C. Abrate, F. Vitale, and F. Bonchi 
    \newline Proceedings of the VLDB Endowment 19 (VLDB 2026)

    \item \label{Pap:mmfp} \textit{Beyond Shortest Paths: Node Fairness in Route Recommendation}~\cite{ferrara2025beyond} 
    \newline \textbf{A. Ferrara}, D. Garc{\'\i}a-Soriano, and F. Bonchi
    \newline Proceedings of the VLDB Endowment 18(9) (VLDB 2025) 
    \item  \label{Pap:barp} \textit{Bias-aware ranking from pairwise comparisons}~\cite{ferrara2024bias}
    \newline \textbf{A. Ferrara}, F. Bonchi, F. Fabbri, F. Karimi, and C. Wagner
    \newline Data Mining and Knowledge Discovery, 38(4) (DAMI 2024) 
    \item \label{Pap:link} \textit{Link recommendations: Their impact on network structure and minorities}~\cite{ferrara2022link}
    \newline \textbf{A. Ferrara}, L. Esp{\'\i}n-Noboa, F. Karimi, and C. Wagner
    \newline Proceedings of the 14\textsuperscript{th} ACM Web Science Conference (WEBSCI~2022)

    \item  \label{Pap:baltor} \textit{Bounded-Abstention Pairwise Learning to Rank}~\cite{ferrara2025bounded}
    \newline \textbf{A. Ferrara}\textsuperscript{*}, A. Pugnana\textsuperscript{*}, F. Bonchi, and S. Ruggieri{\renewcommand{\thefootnote}{\fnsymbol{footnote}}\footnotetext[1]{Equal contribution.}}
    \newline Proceedings of the 32\textsuperscript{nd} ACM SIGKDD Conference on Knowledge Discovery and Data Mining (KDD 2026)
\end{paperlist}

\subsubsection{As Co-author}
\begin{paperlist}[resume=mypapers]
    
    \item \label{Pap:policy} \textit{Policy advice and best practices on bias and fairness in AI}~\cite{alvarez2024policy} 
    \newline J. M. Alvarez, A. B. Colmenarejo, A. Elobaid, S. Fabbrizzi, M. Fahimi, \textbf{A.
Ferrara}, S. Ghodsi, C. Mougan, I. Papageorgiou, P. Reyero, M. Russo, K. M. Scott, L. State, X. Zhao, and S. Ruggieri
    \newline Ethics and Information Technology 26(2) 2024

    \item  \label{Pap:georg}  \textit{Fairness-Aware Ranking Recovery from Pairwise Comparisons}~\cite{ahnert2024fair}
    \newline G. Ahnert, \textbf{A. Ferrara}, and C. Wagner
    \newline Proceedings of the 18\textsuperscript{th} ACM Web Science Conference (WEBSCI~2026)  
    
    \item   \label{Pap:fairMC} \textit{FairMC Fair-Markov chain rank aggregation methods}~\cite{balestra2024fairmc}
    \newline C. Balestra, \textbf{A. Ferrara}, and E. Muller 
    \newline  International Conference on Big Data Analytics and Knowledge Discovery (DaWaK 2024)

\end{paperlist}

\subsection*{Pre-Prints, Under Review and Workshop papers}

\subsubsection{As Co-author}
\begin{paperlist}[resume=mypapers]

    \item  \label{Pap:super} \textit{Super-Resolution of Urban Socioeconomic Indicators via Graph-Based Recommender Systems}~\cite{Nerini2025GNN}
    \newline F. P. Nerini, C. Borile, \textbf{A. Ferrara}, and A. Panisson
    \newline Accepted at WebAndTheCity workshop at The WebConf 2026 

    \item  \label{Pap:et}  \textit{Beyond demographic parity: Redefining equal treatment}~\cite{mougan2023beyond}
    \newline C. Mougan, L. State, \textbf{A. Ferrara}, S. Ruggieri, and S. Staab

\end{paperlist}

\subsection*{Additional Publications of the Author which are not part of the thesis}

\begin{paperlist}[resume=mypapers]

    \item \label{Pap:multi} \textit{A Multidisciplinary Lens of Bias in Hate Speech}~\cite{reyero2023multidisciplinary}
    \newline P. Reyero Lobo, J. Kwarteng, M. Russo, M. Fahimi, K. Scott, \textbf{A.~Ferrara}, I. Sen, and M. Fernandez
    \newline Proceedings of the International Conference on Advances in Social Networks Analysis and Mining (ASONAM 2023)
\end{paperlist}

\section{Roadmap}

The thesis is structured as follows. In \Cref{chap:introduction}, we introduce the foundations and limits of algorithmic fairness, highlighting the necessary transition from simple point estimates to robust fairness testing, and from treating individuals as isolated entities to understanding them within structured systems. In \Cref{chap:background} we introduce  preliminary notations and the related work, while  \Cref{chap:contributions} contains a summary of the contributions of this thesis.

The core technical contributions are divided into four macro themes, from  \Cref{chap:testing} to  \Cref{chap:safety}. In \Cref{chap:testing}, we address testing and auditing for fairness. We introduce \textit{Size-Adaptive Hypothesis Testing for Fairness} (\Cref{Pap:saft}), a unified framework that turns fairness assessment into an evidence-based statistical decision to handle the high variance of small intersectional groups. Following this, we present \textit{Auditing for Demographic Bias in Opaque Rankings} (\Cref{Pap:condor}), which utilizes conditional distance correlation to measure the residual dependence of a ranking on protected attributes within black-box models. Finally, we move beyond disparate outcomes to inspect procedural logic in \textit{Beyond demographic parity: Redefining equal treatment} (\Cref{Pap:et}), introducing Explanation Disparity to detect when models use different reasoning for different demographic groups.

In \Cref{chap:network}, the focus shifts to fairness in network structures, examining how graph topology shapes the distribution of opportunity. In \textit{Beyond Shortest Paths: Node Fairness in Route Recommendation} (\Cref{Pap:mmfp}), we propose the MMFP algorithm to guarantee individual fairness for network nodes by utilizing forward paths. In \textit{Link Recommendations: Their Impact on Network Structure and Minorities} (\Cref{Pap:link}), we then analyze how link recommendation algorithms impact social networks and their structure, demonstrating their tendency to exacerbate popularity bias and reinforce homophily. To bridge the spatial granularity gap, we also introduce a framework for the \textit{Super-Resolution of Urban Socioeconomic Indicators} (\Cref{Pap:super}) using Graph Neural Networks on user-business interaction graphs.

\Cref{chap:ranking} explores fairness in ranking and pairwise comparisons, dealing with the end-to-end pipeline of hierarchical structures. In \textit{Bias-Aware Ranking from Pairwise Comparisons} (\Cref{Pap:barp}), we propose a probabilistic model that estimates individual evaluator biases to recover an unbiased latent ranking. We further address the recovery phase in \textit{Fairness-Aware Ranking Recovery from Pairwise Comparisons} (\Cref{Pap:georg}), evaluating how sampling strategies interact with ranking algorithms to prevent structural invisibility. For rank aggregation, we introduce FairMC (\Cref{Pap:fairMC}), an in-processing Markov Chain method that rescales edge weights to ensure balanced visibility for protected groups.

In \Cref{chap:safety}, we connect these contributions to the reliability and safety of AI deployment. We present BALToR (Bounded-Abstention  Learning to Rank) (\Cref{Pap:baltor}), a framework that allows ranking models to safely abstain from predictions when uncertainty is high without requiring retraining. The thesis concludes with \textit{Policy advice and best practices on bias and fairness in AI} (\Cref{Pap:policy}), synthesizing our technical findings into actionable AI governance, including continuous monitoring and a holistic Bias Management architecture.

We conclude in \Cref{chap:conclusions} by presenting concluding remarks, limitations, and future work of this thesis.

\chapter{Background and Related work}
\label{chap:background}

\begin{displayquote}
\textit{``Life can only be understood backwards; but it must be lived forwards.''} \\
\hspace*{\fill} -- \textup{Søren Kierkegaard} \cite{kierkegaard1967soren}
\end{displayquote}

\section{Definitions of Algorithmic Fairness}
The study of algorithmic fairness emerged at the intersection of machine learning, law, and political philosophy in response to the growing societal impact of automated decision-making systems. Foundational literature has focused on formalizing the ethical concept of ``fairness'' into mathematical definitions, establishing theoretical limitations of these definitions, and identifying the root causes of algorithmic bias.

A central organizing distinction in the algorithmic fairness literature is the difference between individual and group-level fairness metrics.

\subsubsection*{Individual Fairness} Individual fairness  was introduced by Dwork et al.~\cite{dwork2012fairness} in their seminal paper ``Fairness Through Awareness'' and it is based on the Aristotelian principle that ``similar individuals should be treated similarly"~\cite{rowe2002nicomachean}. Mathematically, this is often framed as a Lipschitz condition: for a distance metric $d$ in the feature space and a distance metric $D$ in the outcome space, a mapping $M$ is individually fair if for any individual $x,y$: $D(M(x),M(y))\leq d(x,y)$. However, while conceptually robust, applying individual fairness in practice is often hindered by the inherent difficulty of defining appropriate similarity metrics $d$ and $D$ that correctly capture domain-specific nuances and ethical intuitions. 

\paragraph{Maxmin Fairness} In this thesis, in particular in \Cref{Pap:mmfp}, we consider an alternative framework of  Individual Fairness known as Maxmin Fairness, as defined by García-Soriano and Bonchi \cite{garcia2021maxmin,garcia2020fair}. This approach bypasses the need to explicitly define the distance metrics $d$ and $D$, and instead prioritizes the most disadvantaged individuals. Inspired by John Rawls's theory of distributive justice~\cite{rawls2017theory}, which argues that social inequalities should be arranged to benefit the worst-off, this definition evaluates fairness through an ex-ante probabilistic lens.  Informally, a probability distribution over the set of solutions of a problem is maxmin-fair if it is impossible to improve the satisfaction probability of any
individual without decreasing it for some other individual which is no better off~\cite{garcia2020fair} (where an individual's satisfaction probability is the  probability that they receive a favorable or desired outcome under the chosen randomized distribution over all feasible deterministic solutions).

\subsubsection*{Group Fairness} In contrast to individual fairness, group fairness assesses whether aggregate statistical properties of a model's outputs are equal across predefined demographic groups (often defined by ``sensitive" or ``protected" attributes like race, gender, or age). 

The literature categorizes group fairness into several distinct mathematical criteria, mainly depending on the relationship between the sensitive attribute $A$, the predicted score or label $\hat{Y}$, and the true label $Y$.  A comprehensive list of fairness definitions can be found in Verma et al.~\cite{verma2018fairness}. We recall, here, a few of the most relevant definitions.

\paragraph{Independence (Demographic / Statistical Parity):} 
Requires the prediction to be statistically independent of the sensitive attribute
\[
\hat{Y} \perp A .
\]
It mandates that the rate of positive outcomes (e.g., being hired or receiving a loan) is identical across all demographic groups, regardless of underlying qualification distributions.

\paragraph{Separation (Equalized Odds):} 
Formalized by Hardt et al.~\cite{hardt2016equality}, Equalized Odds requires that the prediction is independent of the sensitive attribute conditional on the true label
\[
\hat{Y} \perp A \mid Y .
\]
Practically, this means the model must achieve equal True Positive Rates (TPR) and equal False Positive Rates (FPR) across all demographic groups.

\paragraph{Sufficiency (Predictive Rate Parity / Calibration):} 
This requires that the true label is independent of the sensitive attribute conditional on the prediction
\[
Y \perp A \mid \hat{Y} .
\]

Perhaps one of the most crucial foundational result in algorithmic fairness is the mathematical incompatibility of certain group fairness metrics. Concurrent and independent work by Kleinberg et al.~\cite{kleinberg2016inherent} and Chouldechova~\cite{chouldechova2017fair}, spurred by the controversy surrounding the COMPAS recidivism risk algorithm, proved what is now known as the \textit{Impossibility Theorem of Fairness}. They demonstrated that if the underlying base rates of the target variable differ between demographic groups, it is impossible to simultaneously satisfy Separation (Equalized Odds) and Sufficiency (Calibration), except in highly restrictive cases. Formally, when
\[
P(Y=1 \mid A=a) \neq P(Y=1 \mid A=b),
\]
no predictor $\hat{Y}$ can satisfy both
\[
\hat{Y} \perp A \mid Y \quad \text{and} \quad 
Y \perp A \mid \hat{Y}
\]
simultaneously.

This inherent trade-off forced the research community to acknowledge that algorithmic fairness cannot be solved by simply finding a ``perfect'' metric; practitioners must explicitly choose which ethical trade-offs to make based on the specific context of the domain.

Furthermore, subsequent foundational work expanded the study of fairness into causal modeling.

\paragraph{Causal Fairness.}
Kusner et al.~\cite{kusner2017counterfactual} introduced the concept of \textit{Counterfactual Fairness}, grounded in Pearl’s causal inference framework~\cite{Pearl2009-PEACII}. A model is considered counterfactually fair if its prediction for an individual remains the same in a counterfactual world in which the individual’s sensitive attribute had taken a different value.

Formally, a predictor $\hat{Y}$ is counterfactually fair if
\[
\hat{Y}_{A \leftarrow a}(U) = \hat{Y}_{A \leftarrow a'}(U)
\]
for all individuals with latent attributes $U$, where $\hat{Y}_{A \leftarrow a}$ denotes the prediction under an intervention setting the sensitive attribute to $a$. Operationalizing this definition requires specifying a causal Directed Acyclic Graph (DAG) that represents the structural relationships between variables and allows the identification of the direct and indirect pathways through which a sensitive attribute influences an outcome.

\section{Detecting Unfairness}

The theoretical formalization of algorithmic fairness into mathematical definitions, such as Statistical Parity, Equalized Odds, and Counterfactual Fairness, establishes first constraints required for equitable machine learning. However, transitioning from theoretical definitions to the practical, operational detection of unfairness in deployed systems represents a profound epistemic and engineering challenge. Detecting unfairness is not merely the act of evaluating a static metric on a hold-out dataset; it requires determining whether observed statistical disparities are artifacts of random sampling variance, reflections of historical systemic prejudices embedded within training data, or the direct mechanical consequences of an algorithm's internal architecture and decision boundaries.

The earliest and most rudimentary approaches to detecting unfairness relied on computing a single point estimate of a chosen fairness metric—such as the difference in selection rates between two demographic groups—and comparing it against a predefined, often arbitrary threshold (for example, the U.S. Equal Employment Opportunity Commission’s traditional ``four-fifths rule''~\cite{greenberg1979analysis}). Typically, these metrics are reported pointwise, and the
decision to deem an observed metric’s value as a potential issue is addressed by defining a threshold:
if the value of the metric is above this threshold, it is considered as a fairness violation, otherwise, it isn't (\cite{ji2020can,ferrara2025size, cherian2024statistical}). 

To address the limitations of fixed thresholds the literature started to consider confidence intervals and hypothesis testing for detecting fairness violations. Confidence Intervals, obtained via Bootstrapping, for example, have been advocated by Besse et al.~\cite{besse2022survey} and Cherian and Candès~\cite{cherian2024statistical}, and are supported by open library tools like \textsc{Fairlearn}~\cite{weerts2023fairlearn}. Del Barrio et al.~\cite{del2019central}, Besse et al.~\cite{besse2022survey}, and Lo et al.~\cite{lo2025bringing} instead used the asymptotic normality of certain fairness metrics to produce hypothesis tests for fairness violations.

Differently from prior works, in this thesis and in particular in \Cref{sec:SAFT}, we propose a general fairness testing methodology that easily applies to all  group fairness metrics based on the confusion matrix and related conditional probabilities. Furthermore, we address statistical reliability in both small and large samples. 

Besides uncertainty and reliability, another key aspect of unfairness detection is the disentanglement of direct and indirect influence on model outcomes. For example, Marx et al.~\cite{marx2019disentangling} use SHAP values to distinguish between the two types of influences. Adler et al.~\cite{adler2018auditing} focuses on detecting indirect influence, in particular, how some features might indirectly influence
outcomes via other, related features. In~\Cref{sec:CONDOR} we show how to combine a conditional and an unconditional independence test to audit for direct and indirect influence in ranking outcomes. Related to our work are hence conditional independence tests such as KCI~\cite{zhang2012kci} and pdCor~\cite{dcorPartial2014}, and the concept of conditional statistical parity~\cite{dwork2012fairness}. Furthermore, strictly related are also the concepts of Structural Causal Models~\cite{Pearl2009-PEACII} and Interventional Fairness~\cite{salimi2019interventional}.

\section{Mitigating Unfairness}

Algorithmic unfairness mitigation techniques are traditionally categorized into three distinct approaches based on when they intervene in the machine learning lifecycle: pre-processing, in-processing, and post-processing.

\paragraph{Pre-processing methods} tackle bias at the very beginning of the pipeline, before the machine learning model is  trained. They work by directly modifying the initial training dataset to remove underlying historical biases, sanitize feature distributions, or eliminate spurious correlations between sensitive attributes and target outcomes. Common techniques include assigning different weights to specific demographic samples~\cite{kamiran2012data} or mathematically adjusting the features to remove disparate impact~\cite{feldman2015certifying}. The main advantage of pre-processing is that it is highly model-agnostic. Once the dataset is debiased, practitioners can use it to train  standard machine learning algorithms. However, the main disadvantage is that altering the underlying data can inadvertently destroy legitimate, highly predictive patterns. Consequently, these methods often lack strong theoretical guarantees regarding how much they will negatively impact the final model's predictive accuracy.

\paragraph{In-processing methods} mitigate bias during the model's training phase. Instead of changing the input data, they modify the learning algorithm itself. This is typically achieved by adding strict fairness constraints or penalty terms directly into the model's objective loss function, forcing the algorithm to learn fair representations. The main advantage of in-processing is that because the model simultaneously optimizes for both predictive accuracy and fairness during optimization, these methods typically achieve the most optimal mathematical trade-off between utility and fairness. Conversely, the main disadvantage is that they are strictly tied to specific model architectures. 

\paragraph{Post-processing methods} operate at the end of the pipeline, after the model has been trained. They do not modify the input data or the algorithm; instead, they treat the trained model as a static opaque model and systematically adjust its final predictions, thresholds, or rankings to ensure the final outputs satisfy certain fairness metrics or fairness constraints. The main advantage of post-processing is that these methods are very versatile and completely avoid the need for expensive, time-consuming model retraining. This makes them the ideal choice for retrofitting fairness into proprietary or legacy systems that are already deployed in production. The main disadvantage is that modifying outputs artificially often disrupts the model's quality and properties. 

\subsubsection*{Unfairness Mitigation in Network Structures}

Traditional algorithmic fairness research has primarily focused on predictive models operating on independent data points. However, many real-world systems involve relational data, where entities interact through network structures. In such settings, outcomes are influenced not only by individual attributes but also by the topology of the network and the dynamics of interaction among nodes. As a result, fairness must be studied not only at the level of individual predictions but also at the level of structural relationships and connectivity patterns.

A central insight from network science is that network topology itself can encode structural inequalities. Individuals occupying peripheral or sparsely connected positions may experience reduced access to information, influence, or opportunities compared to highly connected nodes. Classic work in social network analysis has shown that structural position, captured by measures such as degree centrality~\cite{freeman1978centrality}, betweenness centrality~\cite{freeman1978centrality}, and PageRank~\cite{page1999pagerank}, plays a crucial role in determining visibility and influence within networks. When algorithmic systems rely on such structural features, existing inequalities in connectivity can translate directly into disparities in algorithmic outcomes \cite{bachmann2026network}.

These concerns have motivated a growing body of research on algorithmic fairness in graph-based systems. Recent surveys highlight several core tasks where fairness issues arise, including node classification, link prediction, ranking, and influence maximization~\cite{saxena2024fairsna, dong2023fairness}. In these tasks, structural dependencies violate the independence assumptions underlying many traditional fairness metrics, creating new forms of bias propagation through the network.

One important source of unfairness in networked systems is homophily~\cite{mcpherson2001birds}, the tendency of individuals to connect with others who share similar attributes. Homophily has long been recognized as a fundamental mechanism shaping social networks~\cite{mcpherson2001birds}. While homophily is a natural social phenomenon, it can lead to the formation of tightly connected clusters that isolate minority groups from the rest of the network \cite{karimi2025minorities}. When machine learning models exploit network connectivity, for instance through random walks,  graph embeddings or message passing, these structural patterns can lead to biased representations and discriminatory predictions. Because  models often aggregate information from neighboring nodes, they can inadvertently propagate demographic biases present in the network structure. Several studies have shown that node embeddings learned from biased graphs can encode sensitive attributes even when these attributes are not explicitly provided as input~\cite{bose2019compositional,rahman2019fairwalk}. As a result, downstream tasks built on top of these embeddings may inherit and amplify structural disparities.
To address these issues, a number of fairness-aware graph learning methods have been proposed. One line of work focuses on modifying the random walk sampling process used in graph embeddings. For example, FairWalk~\cite{rahman2019fairwalk} and Crosswalk~\cite{khajehnejad2022crosswalk} alter transition probabilities in random walks to balance visits across demographic groups, thereby reducing bias in learned node representations. Another line of research introduces fairness constraints directly into graph neural network training objectives. Methods such as FairGNN~\cite{dai2022learning} incorporate adversarial learning mechanisms to prevent node embeddings from encoding sensitive attributes, while other approaches introduce regularization terms that enforce group fairness in node classification tasks~\cite{bose2019compositional,ma2021subgroup}. These in-processing techniques aim to jointly optimize predictive performance and fairness during model training.
Beyond representation learning, fairness issues also arise in graph-based ranking algorithms. Ranking methods such as PageRank are widely used to measure node importance and allocate visibility in information networks. However, because these algorithms rely heavily on link structure, they may disproportionately favor nodes that already occupy central positions in the network. To mitigate such disparities, researchers have proposed fairness-aware variants of PageRank and other ranking algorithms that incorporate group-level constraints or adjust teleportation probabilities to ensure more balanced exposure~\cite{tsioutsiouliklis2021fairness,krasanakis2020applying}. 

A further important challenge concerns the dynamic evolution of networks under algorithmic recommendations. In many online platforms, algorithms suggest new connections or interactions between users. These link recommendation systems are typically optimized for predictive accuracy, often reinforcing existing patterns of homophily and preferential attachment. As a result, recommendation algorithms can unintentionally amplify structural inequalities, further isolating minority groups and reinforcing network segregation ~\cite{stoica2018algorithmic,karimi2025minorities}.

Finally, fairness concerns also arise in network processes such as routing and resource allocation, where algorithmic decisions determine how traffic, information, or opportunities flow through the network. Optimization objectives that focus solely on efficiency may produce winner-take-all dynamics, concentrating traffic or visibility along a small set of highly optimized paths while leaving alternative nodes systematically underutilized. These dynamics illustrate how fairness challenges in networks often emerge from the interaction between topology and optimization objectives, rather than from biased predictions alone.
Taken together, this body of work highlights that unfairness in networked systems is fundamentally structural. It arises from the interplay between connectivity patterns, learning algorithms, and optimization goals that govern how visibility and opportunities propagate through the network. Consequently, addressing fairness in these settings requires methods that explicitly account for relational dependencies and, in some cases, intervene directly on the network structure itself.

In this thesis, we explore several aspects of inequalities that may arise in network structures. In particular, we study how unfairness and inequalities can propagate and exacerbate in social networks due to link recommendation algorithms. Furthermore, we propose a method to guarantee individual fairness in route recommendation systems operating on network structures. Finally, we investigate how graph neural networks can be used to better understand socio-economic indicators associated with geographic locations, illustrating how graph-based learning models can provide insights into structural patterns of inequality.

\subsubsection*{Unfairness Mitigation in Hierarchical Structures}
In many algorithmic systems, individuals or items are not evaluated in isolation but are placed within ordered lists that determine access to opportunities such as jobs, loans, university admissions, or visibility in search results. These hierarchical structures translate comparisons into positions, and positions into opportunities. As a consequence, small biases in the comparative process can propagate and crystallize into large disparities in outcomes. 
Unlike classification systems, which produce independent predictions for each individual, ranking systems operate on relative judgments. An item’s position depends not only on its own attributes but also on how it compares with all other candidates. This structural dependence means that fairness cannot be evaluated solely at the level of individual predictions; instead, it must consider the entire comparative process through which merit is inferred, ordered, and aggregated. 

A first source of unfairness arises in the input data used to construct rankings. In many real-world applications, rankings are derived from human judgments or pairwise comparisons, such as peer review processes, crowd-sourced evaluations, hiring assessments, or product ratings. However, extensive research in social psychology and behavioral economics has shown that human evaluators are often influenced by implicit biases, including in-group favoritism and stereotypes affecting perceived competence or quality~\cite{greenwald1995implicit,bertrand2004emily}.
When ranking algorithms treat these judgments as unbiased measurements of latent quality, they effectively conflate prejudice with merit. Preference learning models such as the Bradley–Terry model~\cite{bradley1952rank} or the Plackett–Luce model~\cite{luce1959individual} typically assume that comparisons are generated by rational agents observing a latent utility signal. In practice, however, the observed comparisons may systematically favor certain demographic groups, causing the inferred ranking to reflect social biases rather than true ability.
This issue is particularly pronounced in crowdsourced evaluation settings, where annotators differ in reliability, expertise, and bias. Recent work has therefore explored models that explicitly estimate evaluator bias and reliability in order to disentangle true item quality from systematic judgment distortions~\cite{khetan2016achieving,shah2016permutation,shah2016estimation}. Such approaches aim to recover a more accurate latent ranking by modeling the evaluation process itself rather than assuming unbiased comparisons.

Beyond data collection, unfairness may also arise in the algorithmic mechanisms used to generate rankings. A growing body of research has therefore proposed fairness-aware ranking methods that explicitly incorporate equity constraints into ranking objectives.
One influential line of work focuses on exposure fairness, which studies how visibility in ranked lists is distributed across demographic groups. Because higher-ranked items receive disproportionate attention from users, small positional differences can translate into large differences in exposure or opportunity. Methods such as FA*IR~\cite{zehlike2017fa} and subsequent fairness-aware ranking frameworks aim to guarantee that protected groups receive a minimum level of representation among the top-ranked results.
Another research direction investigates fair ranking under relevance constraints, where algorithms attempt to balance fairness with ranking utility. Approaches such as the fairness-of-exposure framework~\cite{singh2018fairness} model ranking as a constrained optimization problem that redistributes exposure while preserving relevance as much as possible.
Related work has also explored fairness in ranking-based recommender systems, where feedback loops between ranking algorithms and user behavior can amplify disparities over time~\cite{biega2018equity}. These dynamics illustrate how algorithmic hierarchies can interact with user attention patterns to reinforce existing inequalities.

A further challenge arises when multiple rankings must be aggregated into a consensus hierarchy, as in meta-search engines, committee evaluations, or ensemble recommender systems. In such settings, the aggregation mechanism itself can become a source of unfairness.
Many classical rank aggregation algorithms, including the Borda count~\cite{borda1781memoire} and Markov chain–based approaches such as~\cite{dwork2001rank}, combine rankings by repeatedly reinforcing majority preferences. If the input rankings systematically favor certain groups due to position bias or representation bias, the aggregation process can amplify these disparities, producing a consensus ranking that marginalizes protected groups.
Recent research has therefore investigated fairness-aware aggregation mechanisms that adjust transition probabilities or reweight input rankings to prevent the convergence process from reproducing biased structures. These approaches aim to ensure that the final consensus ranking reflects a fair representation of competing candidates rather than merely mirroring the biases embedded in the inputs.

\chapter{Summary of Contributions}
\label{chap:contributions}
\begin{displayquote}
\textit{``The task is not so much to see what no one has yet seen; but to think what nobody has yet thought, about that which everybody sees.''} \\
\hspace*{\fill} -- \textup{Arthur Schopenhauer} \cite{schopenhauer2000parerga}
\end{displayquote}


\section{Contributions to Fairness  Testing and Auditing}

The first stage of the thesis focuses on the \textbf{diagnostic} challenge: how to reliably detect discrimination when data is scarce, models are opaque, or the definition of fairness requires looking beyond simple outcomes.

\spara{Contributions of} \Cref{Pap:saft}: \textbf{\emph{Size-adaptive Hypothesis Testing for Fairness}}. The paper  addresses the statistical brittleness of traditional fairness auditing, which often relies on point estimates and arbitrary thresholds that fail to account for sampling error, especially in small intersectional groups.

\begin{itemize}
    \item \textbf{Methodological:} We introduce Size-Adaptive Fairness Testing (\texttt{SAFT}), a unified hypothesis-testing framework that turns fairness assessment into an evidence-based statistical decision.
    \item \textbf{Theoretical:} For sufficiently large subgroups, we derive Central-Limit results for general group fairness metrics. For smaller, granular intersectional groups, we construct a Bayesian Dirichlet-multinomial estimator to produce reliable Monte-Carlo credible intervals.
    \item \textbf{Empirical:} We provide empirical evidence showing that the Bayesian estimator converges to the theoretical asymptotic behavior. Additionally, we highlight the pitfalls of previously existing fairness measures, especially in the context of intersectional groups.
\end{itemize}  

    
    \spara{Contributions of} \Cref{Pap:condor}: \textbf{\emph{Auditing for Demographic Bias in Opaque Rankings}}. The paper introduces \texttt{Condor} (CONditional Distance cOrrelation for Rankings), a model-agnostic auditing framework to measure the residual dependence of a ranking on protected attributes after accounting for task-relevant features. 
    \begin{itemize}
    \item \textbf{Methodological}: \texttt{Condor} residualizes the ranking and protected attributes within a reproducing kernel Hilbert space and then quantifies the remaining association to provide a conditional independence hypothesis test. 
    \item \textbf{Theoretical} We prove how to combine the conditional test of \texttt{Condor} with an unconditional independence test to achieve a comprehensive causal understanding of the system. The causal interpretation evaluates whether the ranking shows an overall dependence on protected attributes, and whether that dependence remains after adjusting for task-relevant features, allowing auditors to effectively pinpoint the exact nature of the bias. 
    \item \textbf{Empirical:} We validate the framework on synthetic data and semi-synthetic rankings derived from real-world datasets using explicitly controlled parameters for direct and indirect bias. We demonstrate that \texttt{Condor} accurately detects residual demographic dependence regardless of the data's underlying non-linearities.
    \end{itemize}
    
\spara{Contributions of} \Cref{Pap:et}: \textbf{\emph{Beyond demographic parity: Redefining equal treatment}}. The paper moves beyond outcome disparities to inspect the procedural logic of a model. 

\begin{itemize}
    \item \textbf{Conceptual}: Drawing on the principle of ``Equal Treatment,'' we introduce the metric of \textit{Explanation Disparity}.
    \item \textbf{Methodological}: We exploit feature attribution methods (Shapley values) to compare the distributions of explanations across groups, training a meta-classifier to detect when a model relies on different reasoning for different demographic groups, even if the final acceptance rates appear fair.
    \item \textbf{Theoretical}: We demonstrated that while independent explanation distributions guarantee independent prediction distributions, the converse does not hold, meaning Statistical Parity can easily mask disparate treatment. Additionally, we establish the theoretical validity of the Equal Treatment Inspector by proving that under the null hypothesis of statistical independence between the explanations and the protected attribute, the Area Under the Curve of any Classifier Two-Sample Test is exactly $0.5$.
    \item \textbf{Empirical}: We validate the Explanation Disparity metric and the Equal Treatment Inspector on synthetic data and real-world datasets. The experiments confirmed that the inspector accurately detects proxy discrimination and identifies the specific features driving unequal treatment, successfully flagging fairness violations in scenarios where standard metrics like Statistical Parity falsely indicated compliance.
    
\end{itemize}

\section{Contributions to Fairness in Network Structures}

We now shift the focus from individual predictions to \textbf{relational structures}, examining how graph topology and connectivity shape the distribution of opportunity.

\spara{Contributions of} \Cref{Pap:mmfp}: \textbf{\emph{Beyond Shortest Paths: Node Fairness in Route Recommendation}}.

\begin{itemize}

    \item \textbf{Conceptual}: We introduce the novel problem of guaranteeing individual fairness for network nodes in point-to-point route recommendation systems. The problem adopts a Rawlsian notion of maxmin distributional fairness, utilizing randomization to maximize the minimum probability of eligible nodes being included in a recommended route. To achieve this without extreme deviations in travel distance, the paper introduces the concept of \textit{forward paths}, routes where every step strictly decreases the distance to the destination, providing a diverse and fair basis of near-shortest alternatives.

    \item \textbf{Methodological}: We design the \texttt{DAG-FP} algorithm to construct a Directed Acyclic Graph (DAG) that encapsulates all possible forward paths between a source and a destination, effectively avoiding the need to explicitly enumerate a potentially exponential number of routes. We then frame the search for a maxmin-fair probabilistic distribution as a network flow problem on this DAG. This is solved iteratively through the \texttt{MMFP} algorithm using a sequence of small linear programs (LPs).

    \item \textbf{Theoretical}: The research proves that the DAG of forward paths can be constructed in $\Theta(|E| + |V| \log |V|)$ time, matching the asymptotic computational complexity of solving a single shortest-path query via Dijkstra's algorithm. Furthermore, we show that the \texttt{MMFP} algorithm is optimal and we demonstrate that our compact LP formulation requires only polynomially many constraints and variables, ensuring that a maxmin-fair distribution of forward paths can be found in polynomial time. 

    \item \textbf{Empirical}: The framework is tested on several real-world road networks, including massive datasets with millions of edges like the state of Florida and the Eastern part of the USA from OpenStreetMaps. The experiments confirm that the \texttt{MMFP} method achieves a highly equitable distribution of node visits, evidenced by significantly lower Gini coefficients, compared to diversity-based baselines, while keeping path lengths competitive. Additionally, the method proves highly scalable on commodity hardware and outperforms baselines in runtime when generating high volumes of path recommendations for the same query.
     
\end{itemize}

     \spara{Contributions of} \Cref{Pap:link}: \textbf{\emph{Link recommendations: Their impact on network structure and minorities}}. In the paper we investigate people recommender systems in social networks, analyzing how unsupervised link recommendation algorithms (like Personalized PageRank~\cite{jeh2003scaling} and Who-To-Follow~\cite{gupta2013wtf}) influence network evolution. 

     \begin{itemize}
         \item \textbf{Methodological}: We systematically compared five link recommendation algorithms by iteratively applying them to several synthetic bi-populated directed networks 
        to control for varying levels of homophily and minority group sizes. To simulate continuous feedback loops, we added new algorithmic link recommendations and sequentially removed random out-links to maintain constant edge density. Finally, we evaluated the resulting structural changes using the global clustering coefficient, the Gini coefficient of the in-degree distribution, and the visibility of the minority group among the most important nodes.

         \item \textbf{Empirical}: Our experiments demonstrated that all evaluated algorithms increased the clustering coefficient, making the networks more cohesive by closing triangles. However, we also show how these algorithms often exacerbate ``rich-get-richer'' dynamics and reinforce homophily, potentially trapping minority groups in visibility deficits. We highlight \textit{node2vec}~\cite{grover2016node2vec} as a more robust alternative that can mitigate polarization by learning structural roles. 
     \end{itemize}

     \spara{Contributions of} \Cref{Pap:super}: \textbf{\emph{Super-Resolution of Urban Socioeconomic Indicators via Graph-Based Recommender Systems}}. The paper addresses the spatial ``granularity gap'' where coarse data obscures localized structural variations. It introduces a framework that infers high-resolution demographic indicators by using GNNs to learn enriched business representations from user-business interaction graphs. 
    By rendering hidden structural inequalities measurable, this methodology reveals how advantage and disadvantage are distributed across urban topologies.

    \begin{itemize}
        \item \textbf{Methodological}: We propose a GNN framework that models cities as bipartite user-business interaction graphs. The model explicitly enriches business representations with semantic categories and coarse spatial context (postal codes). We introduce a Socioeconomic Super-Resolution pipeline to infer fine-grained attributes (Census Block Groups) using only coarse-grained training labels.
        \item \textbf{Empirical}: Yelp dataset experiments confirm user mobility traces contain strong, quantifiable signals of socioeconomic homophily. Injecting geographic context significantly improves representation quality over standard baselines. The framework successfully bridges the ``granularity gap" by recovering local socioeconomic variations, and the learned embeddings naturally capture spatial neighborhood structures without explicit supervision.
    \end{itemize}

\section{Contributions to Fairness in Ranking and Pairwise Comparisons}

 \Cref{chap:ranking} examines \textbf{hierarchical structures}, specifically the end-to-end pipeline of obtaining and aggregating  pairwise comparisons and rankings.

\spara{Contributions of} \Cref{Pap:barp}: \textbf{\emph{Bias-Aware Ranking from Pairwise Comparisons}}. The paper introduces \texttt{BARP}, a probabilistic model extending the Bradley-Terry framework~\cite{bradley1952rank}. \texttt{BARP} explicitly models and estimates the specific group biases of individual human annotators, allowing for the recovery of a debiased latent ranking from prejudiced pairwise judgments.
\begin{itemize}
    \item  \textbf{Conceptual}: We frame the challenge of biased pairwise comparisons as an evaluator-specific distortion of latent item quality. We recognize that by correcting this distortion directly at the level of individual evaluators, without requiring any group to be pre-designated as protected, the method naturally prevents systemic disadvantages from propagating into the final ranking.
    
    \item \textbf{Methodological}: We develop \texttt{BARP} (Bias-Aware Ranking from Pairwise Comparisons), a novel probabilistic extension of the classic Bradley-Terry model. It introduces individual evaluator bias parameters that distort the perceived scores of items based on their group membership. The framework formulates an explicit log-likelihood function to jointly estimate unbiased item scores and individual evaluator biases via maximum likelihood optimization, while also accommodating multiple, non-binary, and intersectional attributes.
    \item \textbf{Empirical}: We validate the model on both synthetic data with ground-truth labels and real-world datasets. Experiments demonstrate that \texttt{BARP} accurately quantifies hidden individual evaluator biases, achieving a highly accurate reconstruction of the true, unbiased ranking compared to standard and fairness-aware baselines. Furthermore, correcting these evaluation biases naturally resulted in more equalized group exposure in the final ranking.
\end{itemize}

     \spara{Contributions of} \Cref{Pap:georg}: \textbf{\emph{Fairness-Aware Ranking Recovery from Pairwise Comparisons}}. The paper addresses the effects of sampling bias that determines who gets compared in rankings obtained from the pairwise comparisons. 
\begin{itemize}
    \item \textbf{Methodological}: We systematically model the end-to-end ranking pipeline, developing three distinct sampling strategies (Random, Oversampling, and Rank-Based) to simulate representation and popularity biases. We evaluate how these sampling strategies interact with standard (David's Score, RankCentrality, GNNRank) and fairness-aware (Fairness-Aware PageRank) ranking recovery and post-processing algorithms (FA*IR, EPIRA).
    \item \textbf{Empirical}: The findings emphasize that when comparisons are sparse or unevenly sampled, sophisticated graph-based recovery methods are required to prevent structural invisibility from translating into low rankings.
\end{itemize}     
     
    
     \spara{Contributions of} \Cref{Pap:fairMC}: \textbf{\emph{FairMC Fair-Markov Chain Rank Aggregation Methods}}. In the paper, we address the aggregation of multiple rankings into a single consensus ranking. 
\begin{itemize}
    \item \textbf{Methodological}: We develop \texttt{FairMC}, an in-processing rank aggregation algorithm that modifies the underlying directed graph of Markov Chain methods. It rescales edge weights so that the total transition probability to protected nodes equals that to non-protected nodes, ensuring balanced visibility directly within the aggregation mechanism.
    \item \textbf{Empirical}: \texttt{FairMC} outperforms both traditional and state-of-the-art fair aggregation methods in group exposure and top-k fairness. It successfully increased the visibility of protected groups while maintaining an acceptably low Kemeny distance, meaning the final output remained closely aligned with the original input rankings.
\end{itemize}
     

\section{Contributions to Reliability and Safety}

The final chapter connects these technical contributions to the broader requirements of deploying trustworthy AI in high-stakes environments.

     \spara{Contributions of} \Cref{Pap:baltor}: \textbf{\emph{Bounded-Abstention Pairwise Learning to Rank}}. The paper introduces \texttt{BALToR}, a framework that enhances safety by allowing ranking models to abstain from making a prediction when the underlying uncertainty is high. 

\begin{itemize}
    \item \textbf{Theoretical}: We provide a mathematical characterization of the optimal abstention strategy, proving that the optimal selection function relies on thresholding the conditional risk of the ranker based on a target coverage constraint.
    \item \textbf{Methodological}: We develop \texttt{BALToR} (Bounded-Abstention Learning To Rank), a model-agnostic, plug-in algorithm that equips any pre-trained ranking model with a safety deferral mechanism. It operationalizes the theoretical strategy by estimating conditional risk (via prediction confidence or margin) and selectively abstaining from uncertain pairwise comparisons without requiring model retraining.
    \item \textbf{Empirical}: We validate the framework across four benchmark learning-to-rank datasets and demonstrate that \texttt{BALToR} significantly improves ranking accuracy on the accepted pairs, strictly respects the predefined abstention budget (coverage constraint), and evenly distributes rejections across classes to ensure the abstention mechanism itself does not introduce new biases.
    
\end{itemize}

\spara{Contributions of} \Cref{Pap:policy}: \textbf{\emph{Policy advice and best practices on bias and fairness in AI}}. In the paper, we broadly synthesize the technical findings on bias and fairness in AI into best practices.

\begin{itemize}
    \item \textbf{Methodological}: We propose a holistic Bias Management architecture that addresses vulnerabilities across the entire AI lifecycle rather than relying on isolated, post-hoc debiasing interventions. The framework integrates causal reasoning to explicitly model discriminatory mechanisms and champions knowledge-informed AI approaches, which leverage external semantic sources to compensate for the representational gaps and nuances missed by raw observational data.
    \item \textbf{Policy and Practical}: We detail the regulatory friction between EU data protection regimes (GDPR) and non-discrimination mandates, noting that privacy restrictions on sensitive data severely complicate demographic auditing and debiasing. We outline actionable governance practices, including formal AI auditing, documenting the human labor behind data production, implementing continuous monitoring for temporal distribution shifts and feedback loops, and treating Explainable AI with caution due to its susceptibility to instability and secondary biases.
\end{itemize}


\chapter{Testing and Auditing for Fairness}\label{chap:testing}
\begin{displayquote}
\textit{``Doubt is an uncomfortable condition, but certainty is a ridiculous one.''} \\
\hspace*{\fill} -- \textup{Voltaire} \cite{voltaire1919voltaire}
\end{displayquote}

This chapter presents the research on methods for detecting algorithmic unfairness, focusing on the statistical and methodological challenges involved in auditing complex machine learning systems. The works presented in this chapter contribute to \Cref{rq:test} by proposing rigorous frameworks for fairness auditing that move beyond deterministic thresholds toward statistically grounded testing procedures. These approaches aim to improve the reliability of fairness assessments, particularly in scenarios involving small demographic groups, opaque machine learning models, and decision-making pipelines.

In \Cref{sec:SAFT}, we introduce Size-Adaptive Fairness Testing (\texttt{SAFT}), a statistical framework that replaces threshold-based fairness checks with hypothesis testing procedures that account for sampling variability and subgroup size. In \Cref{sec:CONDOR}, we address the challenge of auditing opaque ranking systems, proposing a model-agnostic approach that detects residual demographic dependence in rankings even when the internal logic of the model is inaccessible. Finally, in \Cref{sec:ET}, we move beyond outcome disparities to examine the decision-making process of models, introducing methods that analyze explanation distributions in order to detect cases where models apply different reasoning across demographic groups.

Together, these contributions advance fairness auditing from simple descriptive metrics toward statistically driven diagnostic tools, enabling more reliable assessments of algorithmic systems.

\section{Size-adaptive Hypothesis Testing for Fairness}\label{sec:SAFT}

In this section we introduce Size-Adaptive Fairness Testing (\texttt{SAFT}), a statistical framework designed to improve the reliability of fairness audits. In standard practice, fairness metrics such as statistical parity are computed as single point estimates and compared to predefined thresholds, such as the widely used “four-fifths rule.” However, this deterministic approach ignores the statistical uncertainty of these estimates and treats demographic groups of very different sizes equivalently. As a result, fairness audits may incorrectly flag discrimination in small subgroups due to random fluctuations, or fail to detect significant disparities in larger populations. 

\texttt{SAFT} addresses this limitation by transforming fairness evaluation into a hypothesis testing problem. Instead of relying on arbitrary thresholds, the framework evaluates whether observed disparities are statistically distinguishable from zero, explicitly accounting for the variance of the estimator. The method combines two complementary strategies: an asymptotic test based on the theoretical distribution of fairness metrics, and a Bayesian inference approach that remains reliable when intersectional subgroups are very small and asymptotic assumptions do not hold. 

By providing size-adaptive confidence intervals and hypothesis tests, \texttt{SAFT} enables fairness audits that remain robust across both large and small populations. The framework therefore improves the reliability of fairness diagnosis and contributes to reducing both false positives and false negatives in the detection of discriminatory outcomes.

\subsubsection*{Authors' Contributions}

\noindent
\begin{tabularx}{\textwidth}{@{}l X@{}}
    \toprule
    \textbf{Contribution} & \textbf{Authors} \\
    \midrule
    \textbf{Conceptualization:} & A. Ferrara and all authors \\ 
    \textbf{Writing:} & A. Ferrara and all authors \\ 
    \textbf{Methodology:} & A. Ferrara \\ 
    \textbf{Formal Analysis:} & A. Ferrara \\ 
    \textbf{Code:} & A. Ferrara, F.~Cozzi \\
    \textbf{Experiments:} & F.~Cozzi, A. Ferrara, A. Perotti, A. Panisson \\ 
    \bottomrule
\end{tabularx}

\includepdf[pages=-, width=1.15\textwidth, frame, offset=0 -7, pagecommand={}]{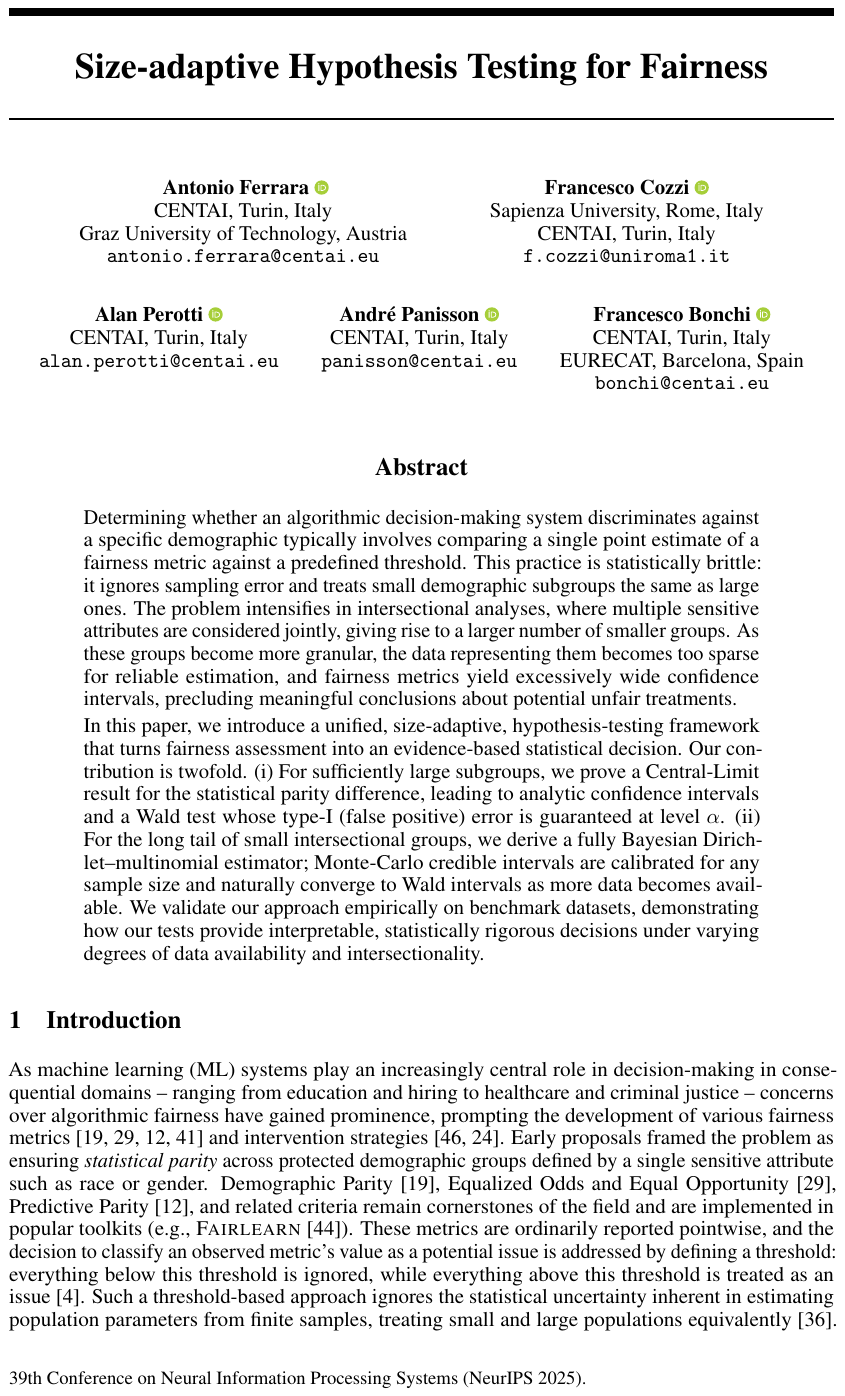}

\section{Auditing For Demographic Bias in Opaque Rankings}\label{sec:CONDOR}

While fairness metrics can be computed directly for classification models, auditing ranking systems presents additional challenges, particularly when the underlying model is opaque or proprietary. In many real-world applications, such as search engines, recommender systems, and hiring platforms, the internal logic of the ranking algorithm may not be accessible to auditors. In such cases, detecting unfairness requires analyzing the observable outputs of the system without relying on knowledge of its internal structure.

This section addresses this challenge by proposing a model-agnostic auditing framework for detecting demographic bias in opaque ranking systems. The central idea is to evaluate whether the ranking produced by the system exhibits residual dependence on protected attributes, after accounting for legitimate task-relevant features. If the ranking remains statistically dependent on demographic attributes even after conditioning on relevant variables, this dependence may indicate the presence of unjustifiable bias.

To operationalize this idea, the proposed framework measures conditional dependence between rankings and protected attributes using statistical tools capable of capturing complex, non-linear relationships. By doing so, the method enables fairness audits that remain applicable even in black-box scenarios where the internal decision-making process of the model cannot be inspected directly.

\subsubsection*{Authors' Contributions}

\noindent
\begin{tabularx}{\textwidth}{@{}l X@{}}
    \toprule
    \textbf{Contribution} & \textbf{Authors} \\
    \midrule
    \textbf{Conceptualization:} & A. Ferrara and all authors \\ 
    \textbf{Writing:} & A. Ferrara and all authors \\ 
    \textbf{Methodology:} & F. Vitale, A. Ferrara \\
    \textbf{Formal Analysis:} & Theorem 1: A. Ferrara; Theorems 2 and 3: F. Vitale \\
    \textbf{Code:} & A. Ferrara, C. Abrate \\
    \textbf{Experiments:} & C. Abrate, A. Ferrara\\ 
    \bottomrule
\end{tabularx}

\includepdf[pages=-,width=1.15\textwidth, frame, offset=0 -7, pagecommand={}]{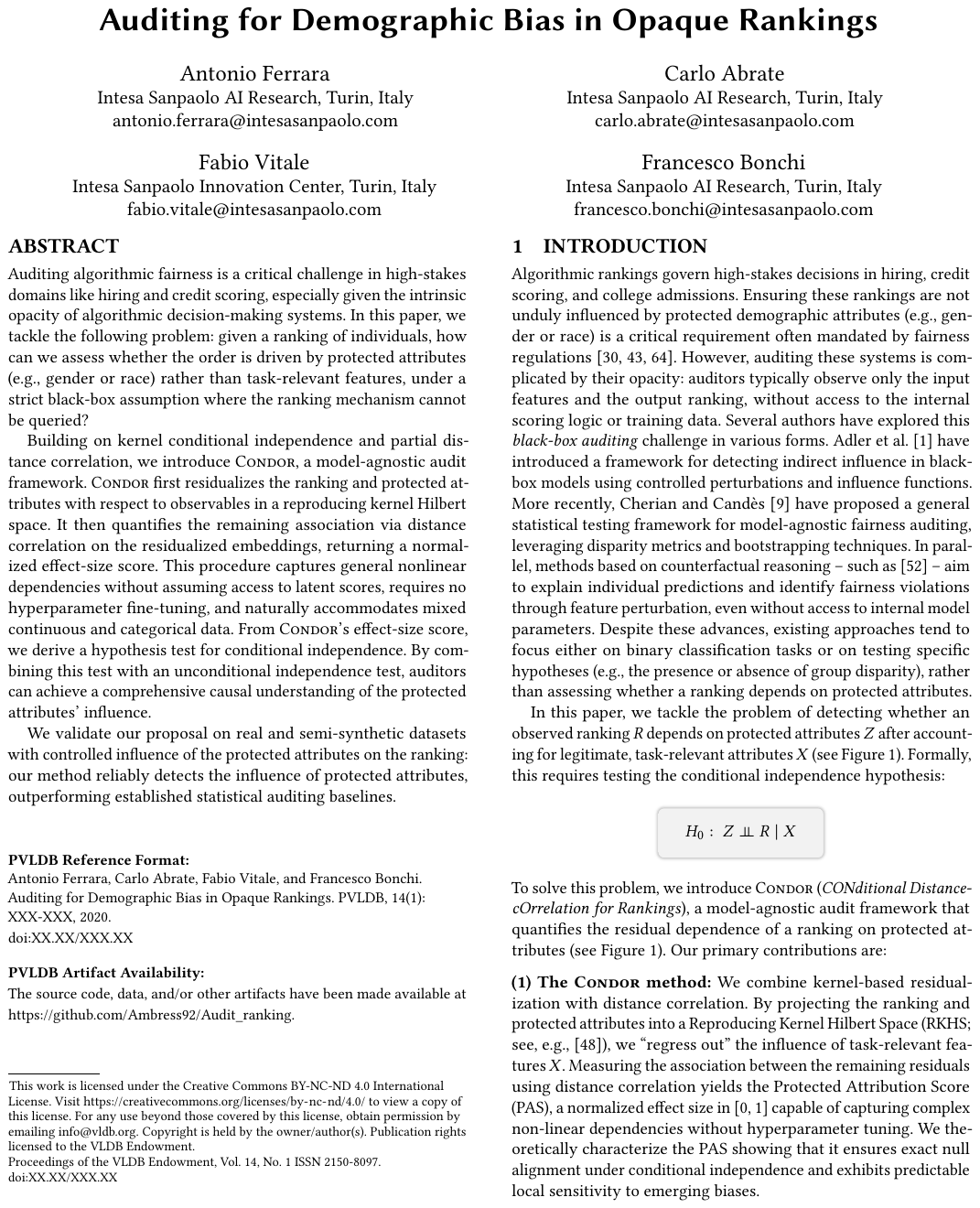}

\section{Beyond Demographic Parity: Redefining Equal Treatment}\label{sec:ET}

Traditional fairness auditing focuses primarily on disparate outcomes, evaluating whether different demographic groups receive favorable decisions at similar rates. However, equal outcomes do not necessarily imply fair treatment. A model may achieve parity in outcomes while still relying on different reasoning or decision-making logic for different groups, for example by using different proxy variables or feature interactions.

The work presented in this section addresses this limitation by introducing a framework for auditing procedural fairness, focusing on how decisions are made rather than only on their final outcomes. The key idea is to analyze the distributions of model explanations or feature attributions across demographic groups. If the explanations that justify decisions differ systematically across groups, this may indicate that the model relies on distinct reasoning processes when evaluating individuals from different demographics.

To detect such disparities, the proposed approach compares explanation distributions and trains a meta-classifier capable of distinguishing whether a given explanation corresponds to one demographic group or another. If the meta-classifier can successfully distinguish between groups, it suggests that the model’s decision-making process differs across groups, revealing a form of disparate treatment that would remain undetected by traditional outcome-based fairness metrics.

By extending fairness auditing to the analysis of explanation distributions, this work highlights the importance of examining the procedural logic of algorithmic systems, complementing traditional metrics focused solely on outcome disparities.

\subsubsection*{Authors' Contributions}

\noindent
\begin{tabularx}{\textwidth}{@{}l X@{}}
    \toprule
    \textbf{Contribution} & \textbf{Authors} \\
    \midrule
    \textbf{Conceptualization:} & C. Mougan, A. Ferrara, and all authors \\ 
    \textbf{Writing:} & C. Mougan and all authors \\ 
    \textbf{Methodology:} & C. Mougan, A. Ferrara, and all authors  \\
    \textbf{Formal Analysis:} & C. Mougan, A. Ferrara, S. Ruggieri \\
    \textbf{Code:} & C. Mougan \\
    \textbf{Experiments:} & C. Mougan\\ 
    \bottomrule
\end{tabularx}

\includepdf[pages=-,width=1.15\textwidth, frame, offset=0 -7, pagecommand={}]{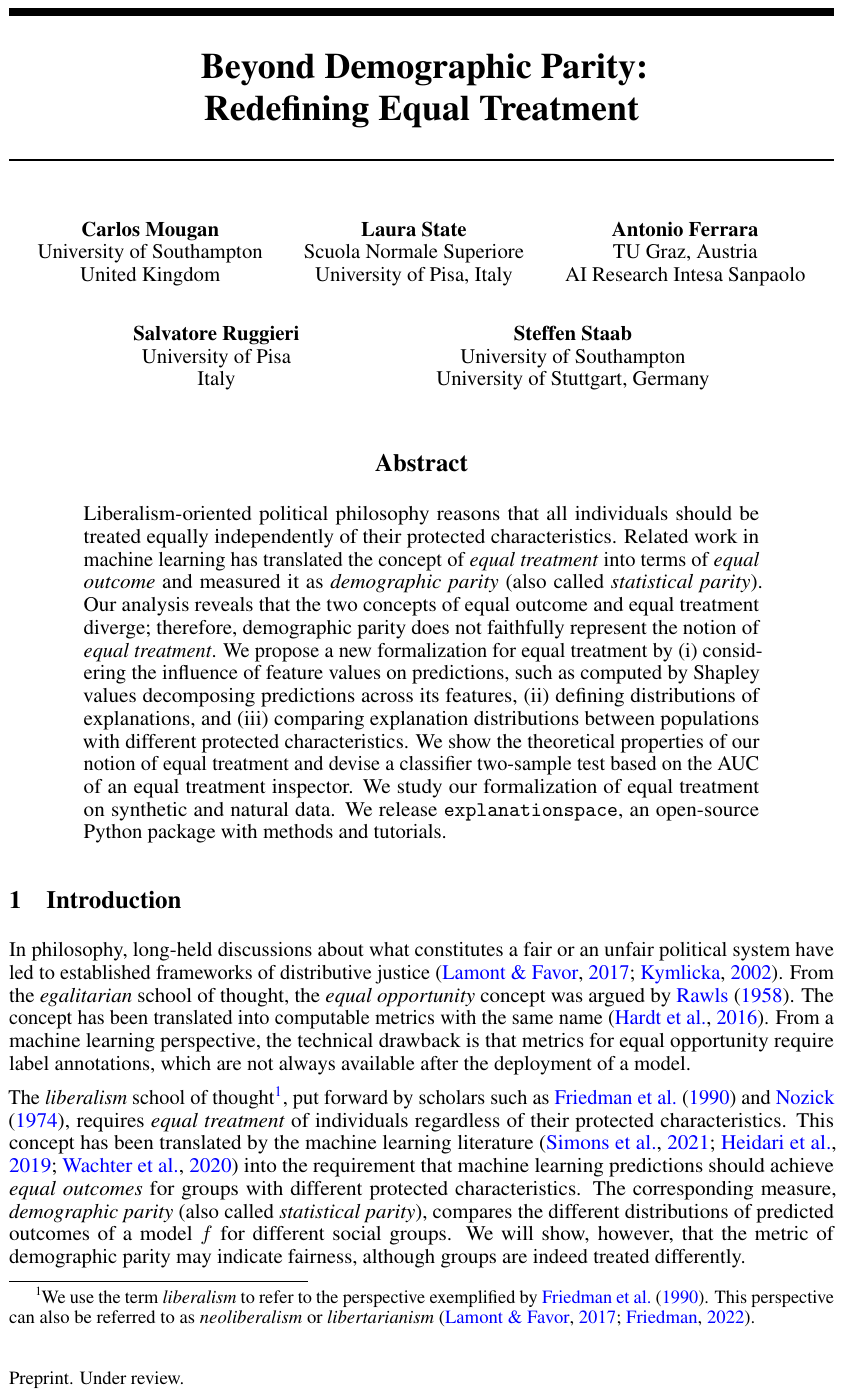}

\chapter{Fairness in Network Structures}\label{chap:network}
\begin{displayquote}
\textit{``When we try to pick out anything by itself, we find it hitched to everything else in the Universe.''} \\
\hspace*{\fill} -- \textup{John Muir} \cite{muir2023my}
\end{displayquote}

This chapter presents the research on algorithmic fairness in networked systems, focusing on how structural properties of networks can influence the distribution of opportunities, visibility, and resources among individuals or entities. Many real-world algorithmic systems operate on relational data, where outcomes are determined not only by individual attributes but also by the topology of interactions and the dynamics of information flow. In such contexts, algorithmic decisions can reinforce or exacerbate structural inequalities embedded in the network.

The publications presented in this chapter contribute to \Cref{rq:networks} by investigating different mechanisms through which unfairness may arise in network structures and by proposing methods to better understand and mitigate such effects. Specifically, the works examine fairness in three distinct but related contexts: route recommendation systems, link recommendation algorithms, and graph-based models for socioeconomic analysis.

In \Cref{sec:MMFP}, we study fairness in route recommendation systems, showing how traditional shortest-path approaches can lead to uneven exposure of nodes within transportation or navigation networks and proposing a fairness-aware alternative. In \Cref{sec:LINK}, we investigate link recommendation algorithms in social networks, analyzing how these systems can influence the evolution of network structure and potentially reinforce inequalities affecting minority groups. Finally, in \Cref{sec:SUPER}, we explore the use of graph-based recommender systems to infer fine-grained socioeconomic indicators in urban environments, illustrating how graph learning techniques can reveal structural patterns in geographic data.

Together, these studies illustrate how fairness concerns arise across different types of network-based algorithms and highlight the importance of incorporating structural considerations into the design and analysis of such systems.

\section{Beyond Shortest paths: Node Fairness in Route Recommendation}\label{sec:MMFP}

Route recommendation systems are widely used in navigation applications and logistics platforms to suggest optimal paths between locations. Most existing systems rely on shortest-path algorithms, which optimize efficiency criteria such as travel time or distance. While effective in improving individual navigation efficiency, these approaches may produce unintended structural consequences when deployed at scale.

In particular, repeatedly recommending the same optimal routes may concentrate traffic or attention on a limited subset of nodes or edges in the network. As a result, certain locations may systematically receive more exposure or benefits, while others remain largely ignored. This raises concerns about fairness at the node level, especially in contexts where exposure to traffic or visitors may have economic or social implications.

The work presented in this section contributes to \Cref{rq:networks} by investigating how fairness considerations can be incorporated into route recommendation systems. Specifically, we introduce a framework for node fairness in route recommendations, which aims to balance traditional efficiency objectives with a more equitable distribution of exposure across nodes in the network. 

\subsubsection*{Authors' Contributions}

\noindent
\begin{tabularx}{\textwidth}{@{}l X@{}}
    \toprule
    \textbf{Contribution} & \textbf{Authors} \\
    \midrule
    \textbf{Conceptualization:} & A. Ferrara and all authors  \\ 
    \textbf{Writing:} & A. Ferrara and all authors \\ 
    \textbf{Methodology:} & A. Ferrara and D. Garcia-Soriano (equally) \\ 
    \textbf{Formal Analysis:} & Prop. 1, 2, 3 and 4: A. Ferrara and D. Garcia-Soriano \newline Lemma 4.1 and Theorem 4.2: D. Garcia-Soriano and A.Ferrara \\ 
    \textbf{Code:} & A. Ferrara \\
    \textbf{Experiments:} &  A. Ferrara \\ 
    \bottomrule
\end{tabularx}

\includepdf[pages=-,width=1.15\textwidth, frame, offset=0 -7, pagecommand={}]{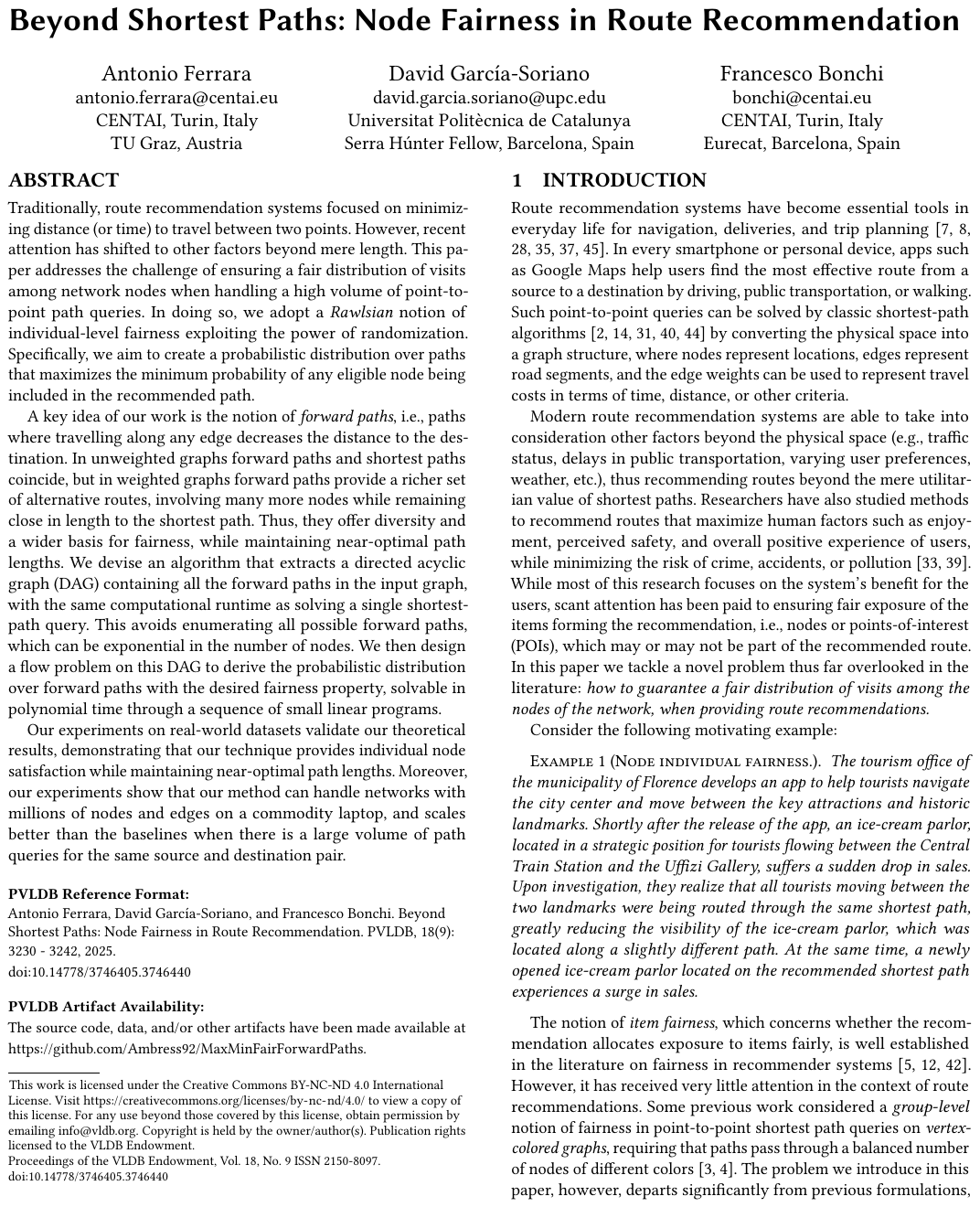}

\section{Link Recommendations: Their Impact on Network Structure and Minorities}\label{sec:LINK}

Link recommendation algorithms play a central role in many online platforms, where they are used to suggest new connections between users. These systems are typically designed to maximize predictive accuracy or user engagement, relying on structural features of the network such as common neighbors, similarity measures, or learned node embeddings.

However, by shaping which connections are formed, link recommendation algorithms can also influence the long-term evolution of the network structure. In particular, they may reinforce existing patterns of homophily or preferential attachment, potentially amplifying structural inequalities affecting minority groups.

The work presented in this section contributes to \Cref{rq:networks} by examining the impact of link recommendation algorithms on the structural properties of social networks. Specifically, we analyze how commonly used recommendation strategies influence network evolution and investigate whether these mechanisms may exacerbate segregation or disadvantage minority groups over time. Through simulation analysis, this study provides insights into the systemic effects of recommendation algorithms, highlighting how seemingly neutral optimization objectives can have unintended consequences for fairness in evolving network structures.

\subsubsection*{Authors' Contributions}

\noindent
\begin{tabularx}{\textwidth}{@{}l X@{}}
    \toprule
    \textbf{Contribution} & \textbf{Authors} \\
    \midrule
    \textbf{Conceptualization:} & A. Ferrara and all authors  \\ 
    \textbf{Writing:} & A. Ferrara and all authors \\ 
    \textbf{Methodology:} & A. Ferrara and all authors \\ 
    \textbf{Code:} & A. Ferrara \\
    \textbf{Experiments:} &  A. Ferrara \\ 
    \bottomrule
\end{tabularx}

\includepdf[pages=-,width=1.15\textwidth, frame, offset=0 -7, pagecommand={}]{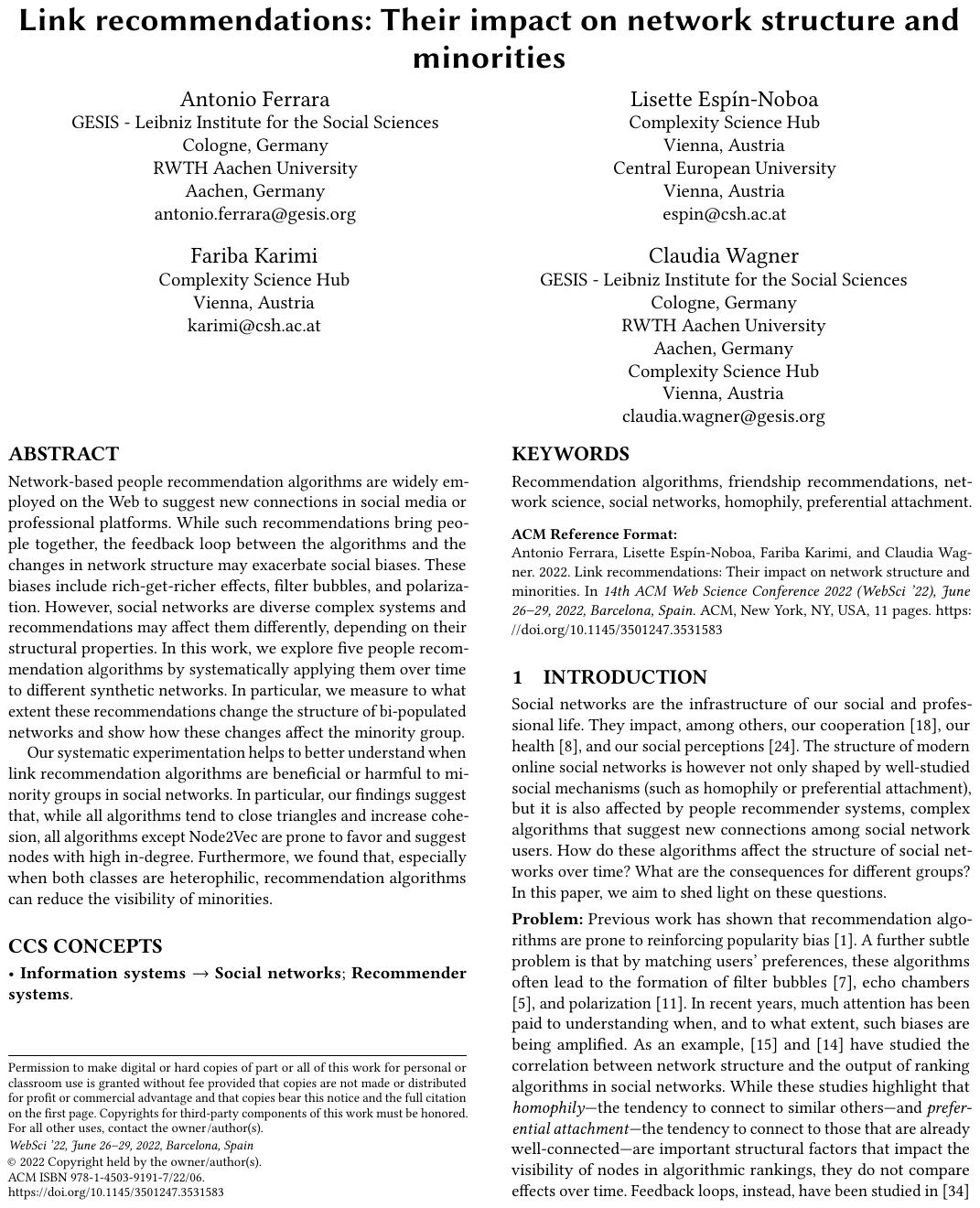}

\section{Super-Resolution of Urban Socioeconomic Indicators via Graph-Based Recommender Systems}\label{sec:SUPER}

Graph-based machine learning methods are increasingly used to analyze complex relational datasets, including those describing urban environments. In cities, many socioeconomic phenomena, such as income distribution, access to services, or mobility patterns, are shaped by spatial and relational dependencies between geographic areas.

However, socioeconomic data are often available only at coarse spatial resolutions, limiting the ability of policymakers and researchers to analyze inequalities at finer geographic scales. Graph-based models provide an opportunity to address this limitation by exploiting structural relationships between locations to infer missing or higher-resolution information.

The work presented in this section contributes to \Cref{rq:networks} by exploring how graph-based recommender systems can be used to infer fine-grained socioeconomic indicators for urban areas. By modeling geographic regions as nodes in a graph and leveraging similarities between them, the proposed approach performs super-resolution of socioeconomic indicators, enabling the estimation of more detailed spatial distributions from limited data. Beyond its methodological contribution, this work illustrates how graph learning techniques can help uncover structural patterns of inequality in urban systems, providing tools for more informed analysis of socioeconomic disparities.

\subsubsection*{Authors' Contributions}

\noindent
\begin{tabularx}{\textwidth}{@{}l X@{}}
    \toprule
    \textbf{Contribution} & \textbf{Authors} \\
    \midrule
    \textbf{Conceptualization:} & F.P. Nerini, A. Ferrara, and all authors  \\ 
    \textbf{Writing:} & F.P. Nerini, A. Ferrara, and all authors  \\ 
    \textbf{Methodology:} & F.P. Nerini, A. Ferrara, and all authors  \\ 
    \textbf{Code:} & F.P. Nerini \\
    \textbf{Experiments:} &  F.P. Nerini \\ 
    \bottomrule
\end{tabularx}

\includepdf[pages=-,width=1.15\textwidth, frame, offset=0 -7, pagecommand={}]{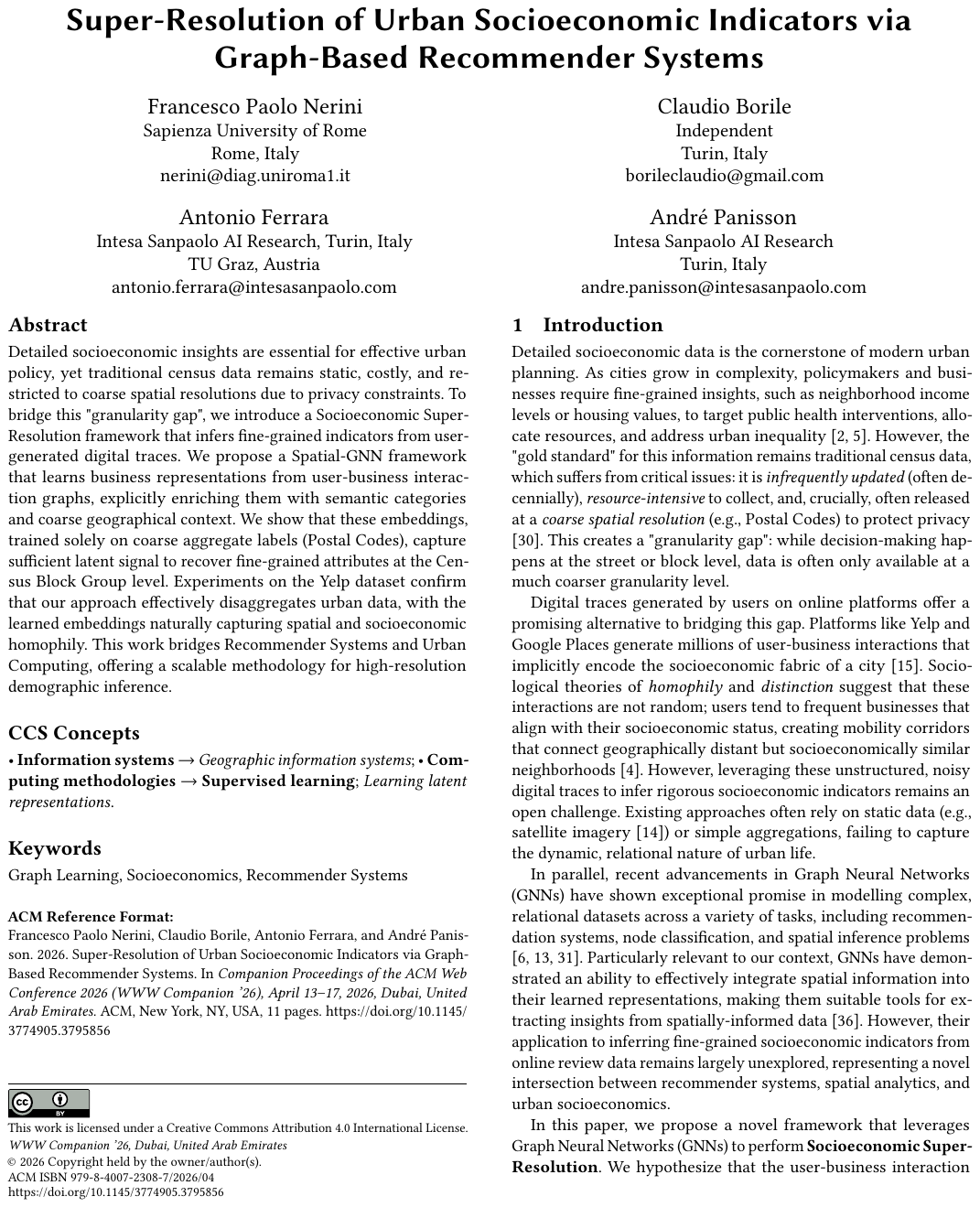}

\chapter{Fairness in Ranking and Pairwise Comparisons}\label{chap:ranking}
\begin{displayquote}
\textit{``In all chaos there is a cosmos, in all disorder a secret order.''} \\
\hspace*{\fill} -- \textup{Carl Jung} \cite{jung2023archetypes}
\end{displayquote}

This chapter investigates fairness in hierarchical systems, focusing on ranking algorithms and pairwise comparison frameworks. While the previous chapter examined fairness in network structures, where outcomes depend on patterns of connectivity, many algorithmic systems instead operate by ordering individuals or items into hierarchical structures. Rankings determine access to opportunities such as employment, funding, visibility in search results, or admission to selective programs. In such contexts, even small distortions in the ranking process can translate into large disparities in exposure and opportunity. 

Unlike classification systems that produce independent predictions for each individual, ranking algorithms operate on relative judgments: the position of an item depends not only on its own attributes but also on how it compares to all other candidates. Consequently, fairness cannot be evaluated solely at the level of individual predictions. Instead, it requires examining the entire pipeline through which merit is inferred, compared, and aggregated into an ordered hierarchy. 

The works presented in this chapter contribute to \Cref{rq:rank} by investigating different sources of unfairness that may arise throughout the ranking pipeline. In particular, the chapter considers three stages of this process: the generation of pairwise comparisons, the recovery of rankings from sampled comparisons, and the aggregation of rankings into a final ordered list.
In \Cref{sec:BARP}, we introduce a probabilistic framework that models evaluator bias in pairwise comparisons in order to recover a more accurate latent ranking. In \Cref{sec:recovery}, we study how the sampling of pairwise comparisons interacts with ranking algorithms, showing how certain sampling strategies may lead to structural invisibility for some items. Finally, in \Cref{sec:FairMC}, we examine the aggregation stage of ranking systems and present a fairness-aware Markov Chain method designed to ensure balanced visibility across demographic groups.
Together, these studies analyze fairness across the entire ranking pipeline, highlighting how biases in human evaluations, data sampling, and aggregation algorithms can propagate through hierarchical systems and ultimately shape the distribution of opportunities.

\section{Bias-Aware Ranking from Pairwise Comparisons}\label{sec:BARP}

Many ranking systems rely on pairwise comparisons as a fundamental source of information. Instead of assigning absolute scores to items, evaluators are asked to compare pairs and indicate which item is preferred. This paradigm appears in a wide range of applications, including peer review, crowdsourced evaluations, hiring processes, and product ratings. Pairwise comparisons are often considered easier and more reliable for human annotators than direct scoring tasks.

However, human judgments are not always unbiased. Extensive research in psychology and behavioral economics has shown that evaluators may be influenced by implicit biases, stereotypes, or favoritism. When ranking algorithms treat these judgments as unbiased measurements of quality, the resulting rankings may inadvertently reflect these biases rather than the true merit of the evaluated items. 

The work presented in this section addresses this challenge by introducing Bias-Aware Ranking from Pairwise Comparisons (\texttt{BARP}), a probabilistic model that explicitly estimates individual evaluator biases. By modeling the evaluation process itself, the method aims to disentangle systematic judgment distortions from the latent quality signal underlying the comparisons. The resulting framework enables the recovery of rankings that better reflect true item quality while accounting for heterogeneity and bias among evaluators.

\subsubsection*{Authors' Contributions}

\noindent
\begin{tabularx}{\textwidth}{@{}l X@{}}
    \toprule
    \textbf{Contribution} & \textbf{Authors} \\
    \midrule
    \textbf{Conceptualization:} & A. Ferrara and all authors \\
    \textbf{Writing:} & A. Ferrara and all authors \\
    \textbf{Methodology:} & A. Ferrara and all authors \\
    \textbf{Formal Analysis:} & A. Ferrara \\
    \textbf{Code:} & A. Ferrara \\
    \textbf{Experiments:} & A. Ferrara \\
    \bottomrule
\end{tabularx}

\includepdf[pages=-,width=1.01\textwidth, frame, offset=0 -7, pagecommand={}]{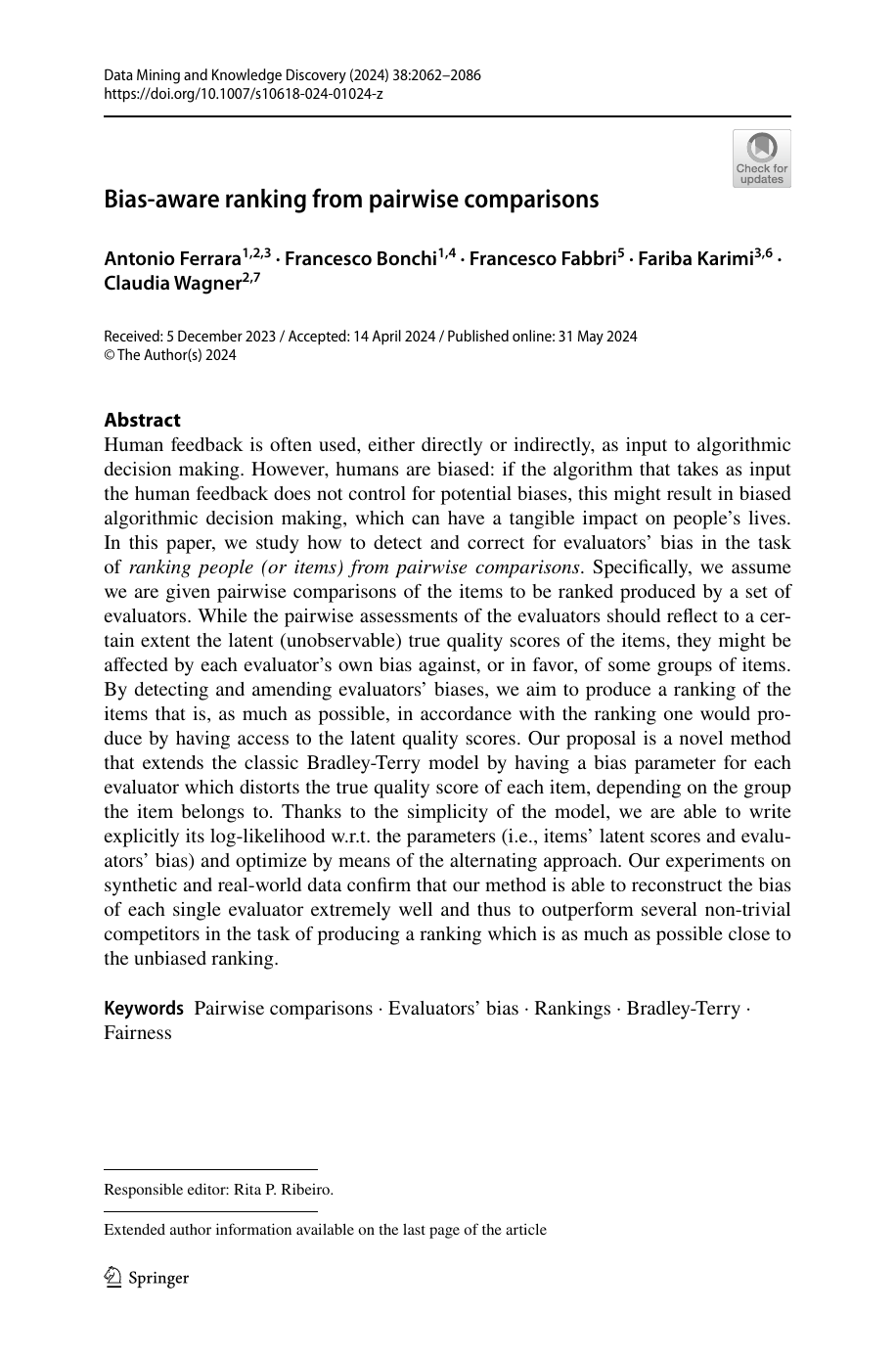}

\section{Fairness-Aware Ranking Recovery from Pairwise Comparisons}\label{sec:recovery}

Even when pairwise comparisons are collected without systematic evaluator bias, the sampling process used to gather comparisons can still introduce structural distortions in the resulting ranking. In many real-world systems, it is infeasible to collect comparisons for every possible pair of items, and algorithms must instead recover rankings from a sparse set of sampled comparisons.

The way these comparisons are sampled can significantly influence the inferred ranking. Some items may receive many comparisons, while others may remain rarely evaluated. As a result, certain candidates may become structurally underrepresented, limiting their chances of appearing in favorable ranking positions regardless of their true quality.

The work presented in this section studies the interaction between sampling strategies and ranking recovery algorithms. In particular, it investigates how different sampling mechanisms affect the visibility and representation of items in the resulting ranking.

\subsubsection*{Authors' Contributions}

\noindent
\begin{tabularx}{\textwidth}{@{}l X@{}}
    \toprule
    \textbf{Contribution} & \textbf{Authors} \\
    \midrule
    \textbf{Conceptualization:} & G. Ahnert, A. Ferrara, C. Wagner \\
    \textbf{Writing:} & G. Ahnert, A. Ferrara, C. Wagner \\
    \textbf{Methodology:} & G. Ahnert, A. Ferrara, C. Wagner \\
    \textbf{Code:} & G. Ahnert, A. Ferrara \\
    \textbf{Experiments:} & G. Ahnert \\
    \bottomrule
\end{tabularx}

\includepdf[pages=-,width=1.15\textwidth, frame, offset=0 -7, pagecommand={}]{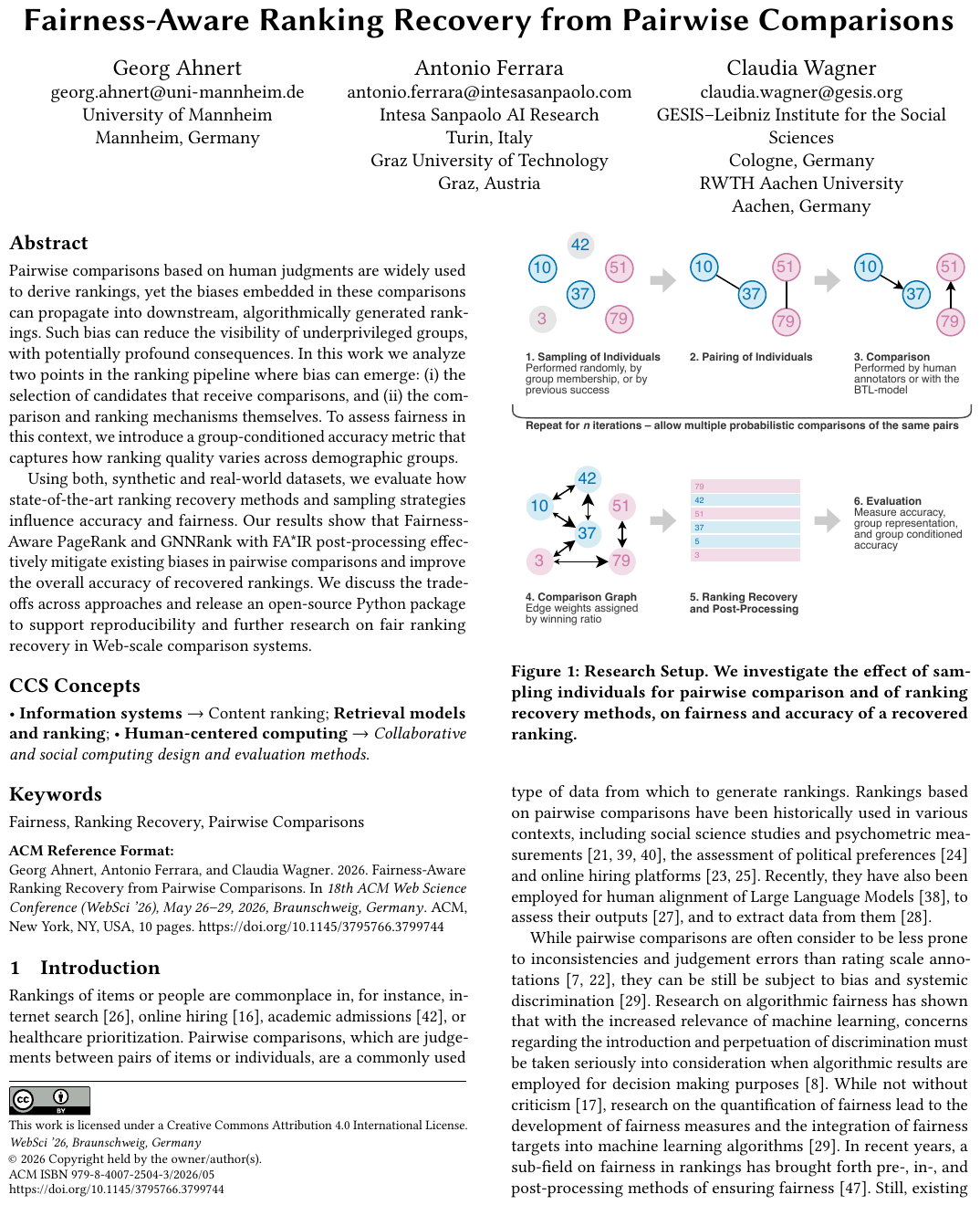}

\section{FairMC Fair-Markov Chain Rank Aggregation Methods}\label{sec:FairMC}

In many applications, rankings are not generated directly from raw data but instead emerge from the aggregation of multiple partial rankings or comparison signals. For example, search engines combine signals from various sources, recommender systems aggregate user feedback, and crowdsourcing platforms merge the judgments of multiple annotators.

Markov Chain–based methods are widely used for this purpose, as they model ranking as a stochastic process over pairwise preferences. However, when the input comparisons or rankings contain structural biases, standard aggregation methods may amplify these disparities, producing final rankings that systematically disadvantage certain groups.

The work presented in this section introduces \texttt{FairMC}, a fairness-aware rank aggregation method based on Markov Chains. The approach modifies the transition dynamics of the aggregation process by rescaling edge weights, enabling the algorithm to balance visibility across demographic groups while preserving the overall structure of the ranking signal. By incorporating fairness considerations directly into the aggregation procedure, the method provides a mechanism for mitigating disparities in the final ranking outcomes.

\subsubsection*{Authors' Contributions}

\noindent
\begin{tabularx}{\textwidth}{@{}l X@{}}
    \toprule
    \textbf{Contribution} & \textbf{Authors} \\
    \midrule
    \textbf{Conceptualization:} & A. Ferrara, C. Balestra \\
    \textbf{Writing:} & C. Balestra, A. Ferrara \\
    \textbf{Methodology:} & A. Ferrara, C. Balestra \\
    \textbf{Code:} & C. Balestra, A. Ferrara \\
    \textbf{Experiments:} & C. Balestra, A. Ferrara \\
    \bottomrule
\end{tabularx}

\includepdf[pages=-,width=1.01\textwidth, frame, offset=0 -7, pagecommand={}]{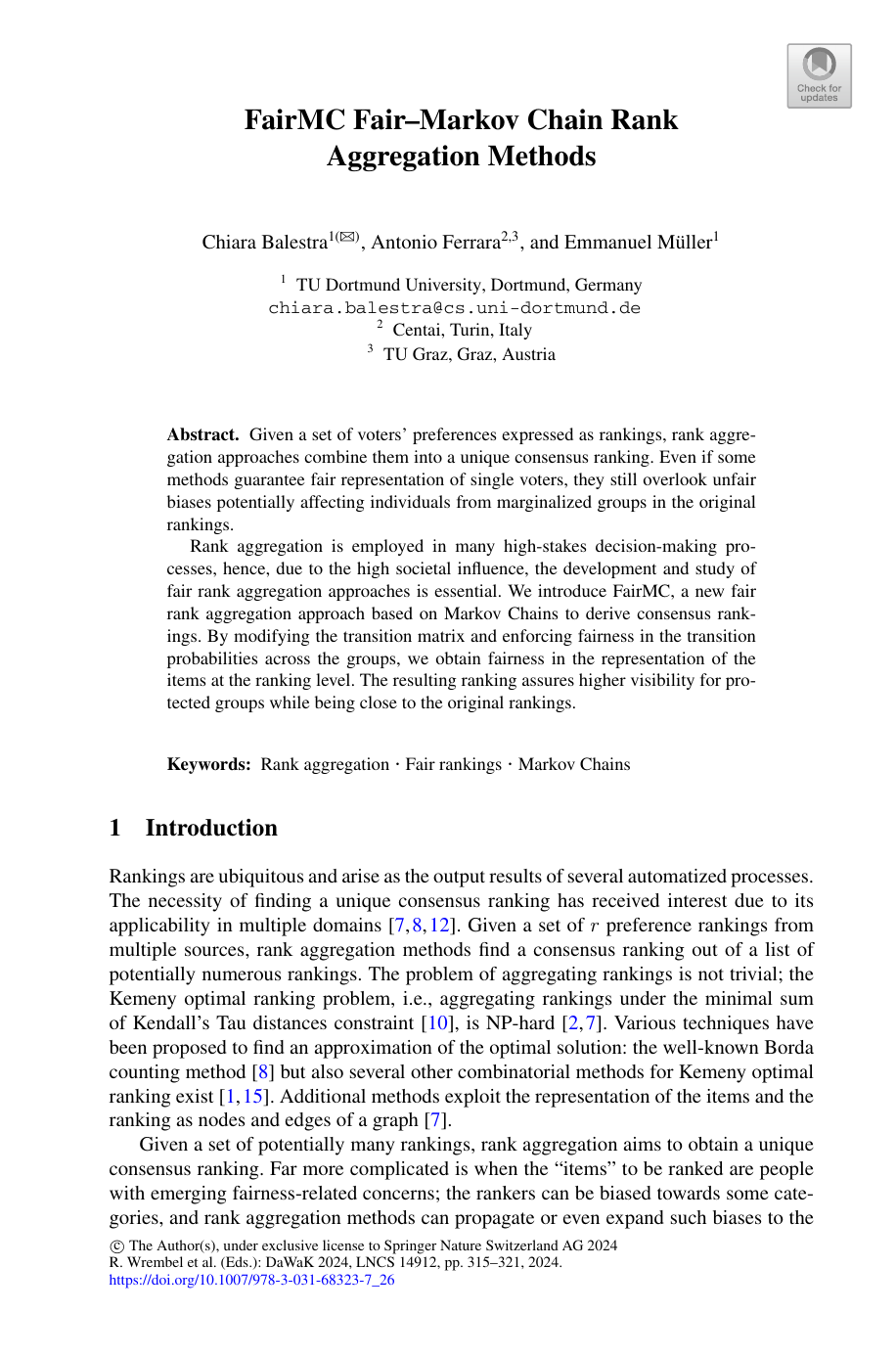}

\chapter{Safety and Policy for Fairness}\label{chap:safety}
\begin{displayquote}
\textit{``And now here is my secret, a very simple secret: It is only with the heart that one can see rightly; what is essential is invisible to the eye.''} \\
\hspace*{\fill} -- \textup{Antoine de Saint-Exupéry} \cite{saintexupery1943little}
\end{displayquote}

This chapter focuses on the reliability, safety, and policy of algorithmic decision-making systems, addressing the challenges that arise when such systems are deployed in high-stakes, real-world environments. While the previous chapters examined how to detect unfairness and how structural inequalities emerge in networked and hierarchical systems, these analyses alone are not sufficient to ensure that AI systems behave safely in practice. In real-world deployments, models must operate under uncertainty, interact with incomplete or biased data, and remain robust to distributional shifts and opaque decision-making processes.

A key limitation of current approaches is that they often assume that models should always produce a prediction. However, in many high-stakes domains, such as healthcare, hiring, or public policy, forcing a decision under uncertainty can lead to harmful outcomes. As a result, ensuring safety requires mechanisms that allow models to recognize their own uncertainty and defer decisions when necessary, as well as broader governance frameworks that address risks across the entire lifecycle of AI systems. 

The works presented in this chapter contribute to \Cref{rq:safety} by proposing both algorithmic and systemic approaches to reliability and safety. Rather than focusing solely on fairness metrics or structural properties, this chapter examines how to design systems that remain robust, trustworthy, and accountable under real-world conditions.

In \Cref{sec:baltor}, we introduce a method for incorporating abstention mechanisms into ranking systems, enabling models to defer uncertain decisions in a principled and controlled manner. In \Cref{sec:policy} we extend the perspective beyond individual algorithms, synthesizing the technical insights of this thesis into policy recommendations and best practices for managing bias and ensuring reliability across the AI lifecycle.

Together, these contributions connect fairness research with the broader goal of trustworthy AI, highlighting that achieving fairness in isolation is not sufficient without mechanisms that ensure safe and reliable deployment.

\section{Bounded-Abstention Pairwise Learning to Rank}\label{sec:baltor}

Ranking systems are increasingly used in high-stakes decision-making contexts, where outcomes depend directly on an individual’s position in an ordered list. In such settings, errors can have significant social and economic consequences. Despite this, most ranking models are designed to always produce a prediction, even when the underlying data is noisy or the model is uncertain.

This work addresses this limitation by introducing Bounded-Abstention Pairwise Learning to Rank (\texttt{BALToR}), a framework that equips ranking models with a safety mechanism based on abstention. Instead of forcing a potentially unreliable decision, the model is allowed to defer predictions when its confidence is low, enabling human intervention or the collection of additional information. 

The approach is grounded in a theoretical characterization of the optimal abstention strategy, which is based on thresholding the conditional risk of the ranking model under a predefined coverage constraint. Building on this result, \texttt{BALToR} provides a model-agnostic, plug-in method that can be applied to existing ranking systems without requiring retraining. 

By selectively abstaining on uncertain comparisons, the method improves the reliability of ranking outcomes while ensuring that the abstention mechanism itself does not introduce additional biases. This work highlights the importance of moving beyond models that always produce a prediction toward selective prediction frameworks, where models can explicitly manage uncertainty as part of their decision-making process.

\subsubsection*{Authors' Contributions}

\noindent
\begin{tabularx}{\textwidth}{@{}l X@{}}
    \toprule
    \textbf{Contribution} & \textbf{Authors} \\
    \midrule
    \textbf{Conceptualization:} & A. Ferrara and all authors \\
    \textbf{Writing:} & A. Ferrara and all authors \\
    \textbf{Methodology:} & A. Ferrara, A. Pugnana \\
    \textbf{Formal Analysis:} & A. Pugnana, A. Ferrara \\
    \textbf{Code:} & A. Pugnana, A. Ferrara \\
    \textbf{Experiments:} & A. Pugnana, A. Ferrara \\
    \bottomrule
\end{tabularx}

\includepdf[pages=-,width=1.15\textwidth, frame, offset=0 -7, pagecommand={}]{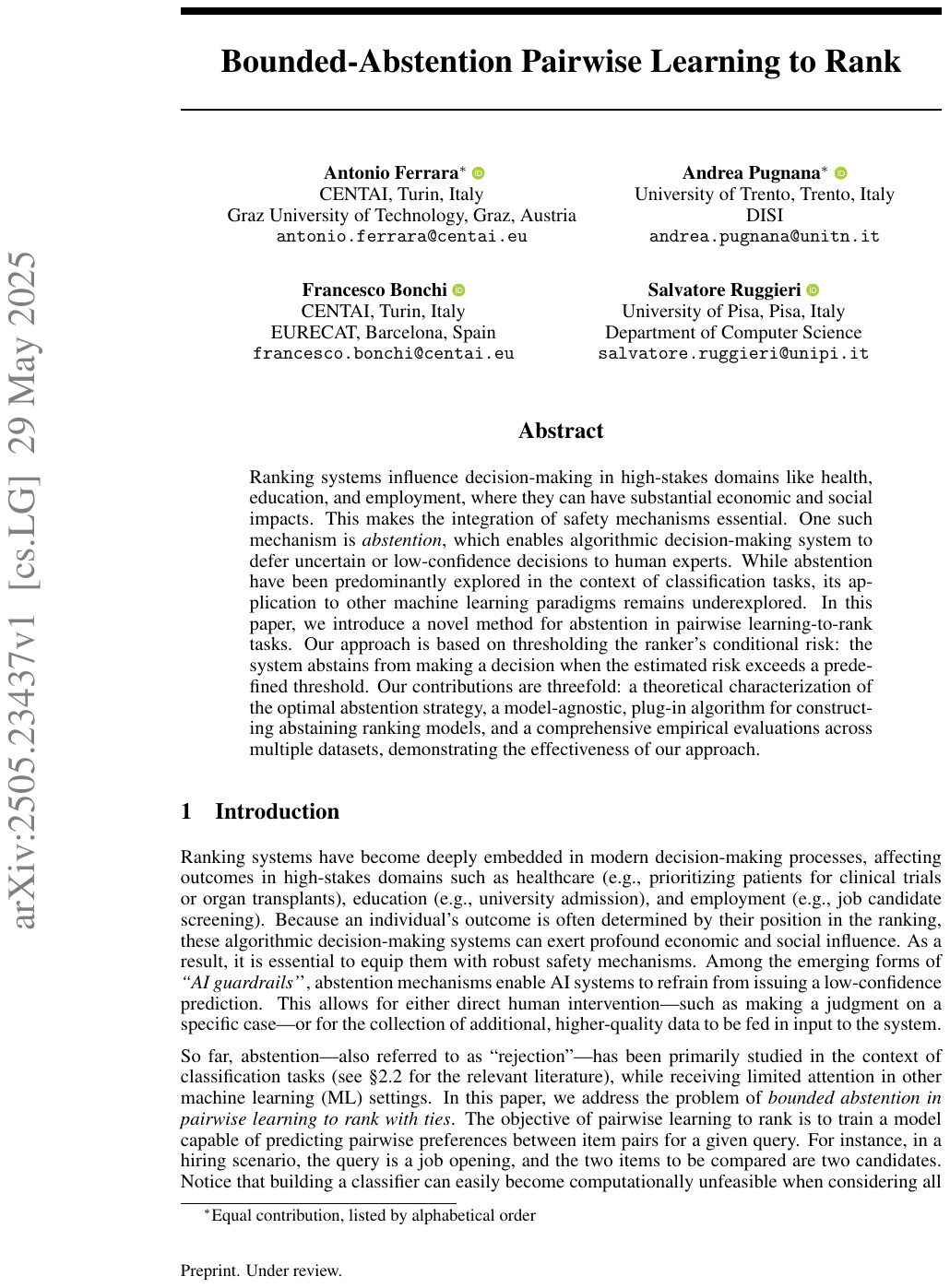}

\section{Policy Advice and Best Practices on Bias and Fairness in AI}\label{sec:policy}

While algorithmic interventions can improve fairness and reliability at the model level, ensuring trustworthy AI requires a broader perspective that considers the entire lifecycle of AI systems. Bias and unreliability can emerge at multiple stages, including data collection, model development, deployment, and ongoing use. Addressing these challenges therefore requires not only technical solutions but also organizational and regulatory practices.

This work synthesizes the technical contributions of the thesis into a set of policy recommendations and best practices for managing bias and ensuring reliability in AI systems. Central to this approach is a holistic bias management framework, which emphasizes that fairness cannot be achieved through isolated, post-hoc corrections but must instead be addressed systematically across all stages of the pipeline. 

The framework highlights several key challenges. First, regulatory constraints, such as those imposed by data protection laws, can limit access to sensitive attributes, complicating fairness auditing and mitigation efforts. Second, AI systems are subject to temporal dynamics, including distribution shifts and feedback loops, which can degrade performance and fairness over time. Third, tools such as explainable AI, while useful, may introduce additional instability or biases if not applied carefully.

To address these issues, the work outlines practical recommendations, including the adoption of continuous monitoring, the integration of causal reasoning to better understand discriminatory mechanisms, and the explicit documentation of data generation processes and human involvement. These practices aim to ensure that AI systems remain reliable, auditable, and aligned with societal and regulatory requirements.

\subsubsection*{Authors' Contributions}

\noindent
\begin{tabularx}{\textwidth}{@{}l X@{}}
    \toprule
    \textbf{Contribution} & \textbf{Authors} \\
    \midrule
    \textbf{Conceptualization:} & J.M. Alvarez and all authors \\
    \textbf{Writing:} & J.M. Alvarez and all authors \newline A. Ferrara mainly wrote the Mitigating bias section \\
    \textbf{Methodology:} & J.M. Alvarez and all authors \\
    \bottomrule
\end{tabularx}

\includepdf[pages=-,width=1.15\textwidth, frame, offset=0 -7, pagecommand={}]{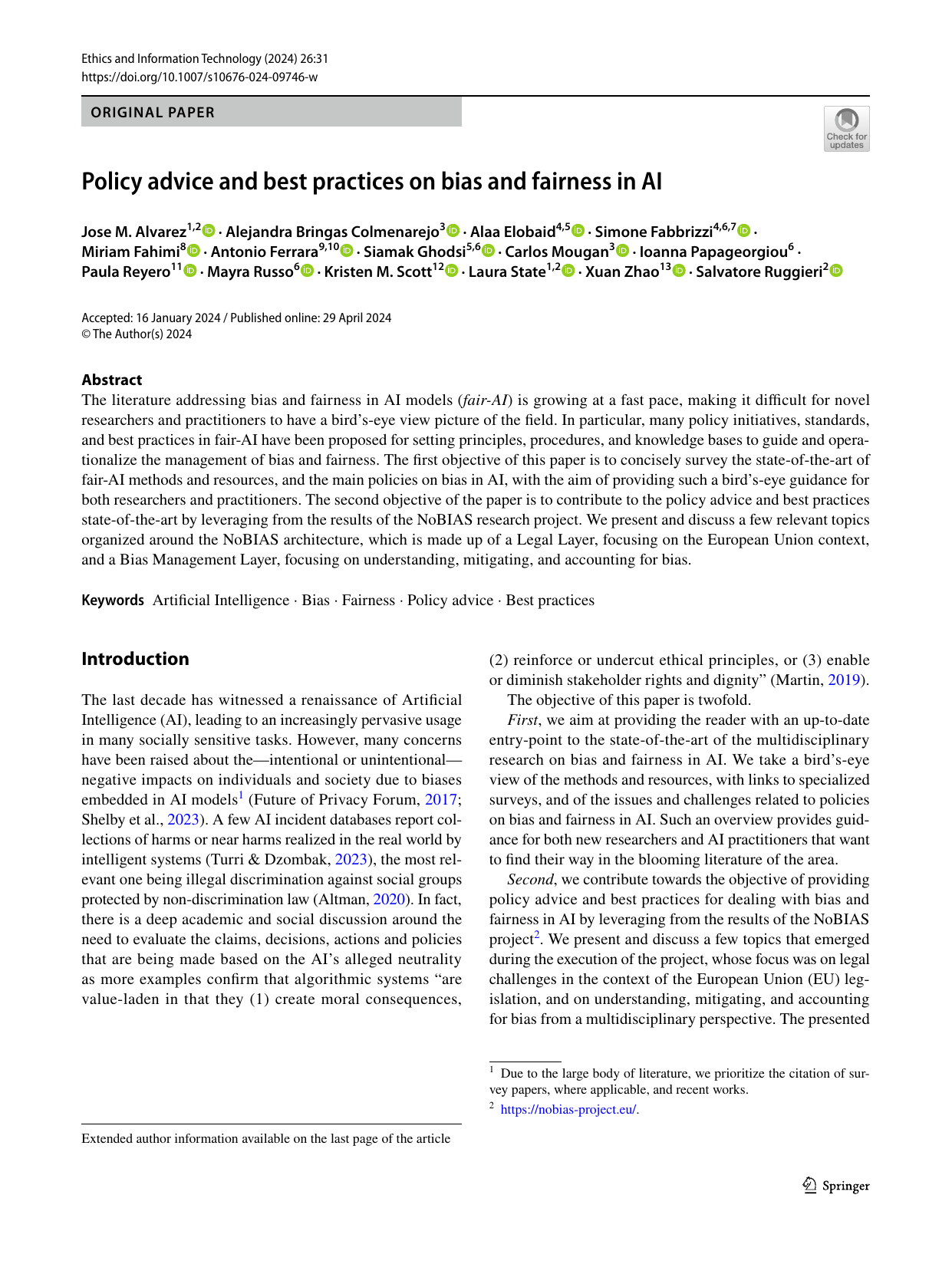}

\chapter{Conclusions, Limitations and Future Work}
\label{chap:conclusions}

\markright{Conclusions, Limitations and Future Work}
\renewcommand{\thesection}{\thechapter}

\begin{displayquote}
\textit{``All that you touch\\ You Change.\\ All that you Change\\ Changes you.\\ The only lasting truth\\ is Change.''} \\
\hspace*{\fill} -- \textup{Octavia E. Butler} \cite{butler1993parable}
\end{displayquote}

\section*{Conclusions}

This thesis argued that algorithmic fairness must evolve along two fundamental dimensions: it must become statistically robust and structurally aware. As algorithmic systems increasingly function as institutional gatekeepers, fairness can neither be assessed through deterministic point estimates nor ensured by adjusting outputs alone. It must be grounded in reliable inference and embedded into the relational mechanisms that govern how opportunities are distributed.
On the diagnostic side, this work reframed fairness auditing as a problem of statistical and causal inference. Instead of relying on brittle point estimates, it introduced size-adaptive testing procedures that account for sampling variability in intersectional groups, conditional independence tests that isolate residual bias in opaque ranking systems, and explanation-based analyses that detect disparate treatment even when outcome distributions appear equal. Together, these contributions move fairness evaluation from threshold comparison to evidence-based decision-making, addressing statistical reliability, causal validity, and procedural neutrality in a unified framework.
On the structural side, the thesis demonstrated that unfairness often emerges from topology and interaction rather than isolated predictions. In routing systems, strict efficiency objectives can structurally concentrate exposure; in link recommendation, feedback loops amplify homophily and popularity bias; in urban inference, coarse measurements obscure localized inequalities; and in ranking pipelines, evaluator bias, skewed sampling, and aggregation dynamics can crystallize prejudice into hierarchical disadvantage. Across these domains, one of the central insights is that fairness must intervene in the inputs, mechanisms, and connectivity patterns that shape visibility and comparison, not only in final outcomes.
Finally, the thesis connected fairness to reliability and governance. By introducing abstention mechanisms for ranking systems and articulating a holistic bias-management architecture, it emphasized that trustworthy AI requires continuous monitoring, explicit modeling of uncertainty, and lifecycle-wide oversight.
In sum, this work advances a comprehensive view of algorithmic fairness as a systemic property: statistically grounded, structurally informed, and operationally embedded. Ensuring fairness in modern AI systems is not a matter of satisfying isolated metrics, but of designing and auditing the institutional processes through which algorithms shape opportunity.

\section*{Limitations} 

In this thesis, we acknowledge that theoretical definitions of algorithmic fairness rely on mathematical abstractions that cannot fully capture the highly complex nature of political philosophy and distributive justice. Consequently, deploying a mathematically fair algorithmic technique does not automatically guarantee fairness within the broader, complex socio-technical system it inhabits. This disconnect between mathematical abstraction and social reality is perhaps most evident in the pervasive necessity of treating human identity as discrete, fixed categories. Whether defining groups for fairness testing or mitigation, these models rely on reducing complex, fluid, and continuous human identities into rigid bins, such as binary gender or predefined racial categories. This epistemological reduction inherently fails to capture the true sociological nuances of identity and marginalization.

Beyond the categorization of identity, this reductionist approach also extends to the evaluation of individuals, leading to a ``ground truth'' illusion regarding merit. Several structural interventions in this thesis, particularly in recovering fair rankings from pairwise comparisons or aggregating crowd-sourced judgments, rely on modeling and isolating human bias to uncover a candidate's latent quality. However, in complex socio-technical systems, ground truth or merit is often inherently subjective, socially constructed, or historically distorted. While mathematical models can elegantly correct for observed statistical deviations, they cannot entirely resolve the philosophical ambiguity of what constitutes an objectively correct ranking.

Even if individual identities and subjective merit could be perfectly quantified, the mathematical models used to process them face their own systemic boundaries. The structural interventions proposed in this thesis, whether redistributing exposure in physical routing networks, mitigating homophily in link recommendations, or aggregating hierarchical rankings, implicitly model these platforms as closed, self-contained environments. In reality, users and opportunities exist within interconnected, open-world ecosystems. A limitation of these structural approaches is that mitigating bias on isolated graphs or ranking systems does not account for how advantages and disadvantages cascade across other platforms, or how users simultaneously use different, potentially biased systems.

Finally, compounding these theoretical and structural limitations are the practical barriers encountered during real-world deployment. The practical implementation of demographic auditing and debiasing frameworks is often severely complicated by real-world privacy restrictions. Data protection regimes, such as the GDPR, strictly limit access to the exact sensitive attributes needed to effectively run these statistical tests and monitor compliance. This creates an ongoing regulatory friction between privacy mandates and non-discrimination goals, presenting a formidable final hurdle to comprehensive AI auditing.

\section*{Future Work}

Just as foundational methods in statistics, optimization, and linear algebra were developed long before deep learning could build upon them, this thesis aims to establish rigorous statistical and structural foundations for algorithmic fairness. Several frameworks introduced in the thesis, including hypothesis testing for fairness auditing, causal diagnostics for opaque systems, and bias-aware preference modeling, are deliberately general: they do not depend on a specific model architecture or application domain. This generality positions them to be extended to the next generation of AI systems, where the challenges of fairness are amplified rather than resolved. In what follows, we outline several broad directions along which this work could be developed further.

On the diagnostic side, the hypothesis-testing frameworks presented here operate mostly in a static setting. Deployed systems, however, require continuous monitoring as new data arrives and distributions shift. Extending size-adaptive fairness testing to a sequential setting, where statistical guarantees must be maintained over time while adapting to distributional drift, would provide the formal underpinning for the continuous monitoring advocated in the policy chapter of this thesis. Furthermore, the resolution limit introduced in this thesis raises deeper information-theoretic questions about the minimum amount of data required to certify a system as fair, and how such minima should inform regulatory standards for deployment.

On the structural side, this thesis studies fairness in routing, link recommendation, and socioeconomic inference as separate problems on distinct networks. In reality, these processes coexist on overlapping topologies: the same urban graph supports both traffic flow and economic activity, the same social network mediates both information propagation and access to opportunity. Understanding how a fairness intervention in one process affects the others requires a unified multi-process perspective that this thesis does not yet provide. 

Regarding the bias-aware preference modeling developed in this thesis, a direct frontier is its extension to the alignment of large language models. Modern alignment pipelines such as Reinforcement Learning from Human Feedback \cite{christiano2017deep} train reward models from human pairwise preferences, a setting that directly instantiates the problem addressed by \texttt{BARP}. Current approaches assume that annotators are unbiased sensors of helpfulness, yet evaluators carry cultural biases and favoritism, precisely the distortions that \texttt{BARP} models through its evaluator-specific bias parameters. Integrating bias-aware preference modeling into reward model training could prevent demographic prejudice from contaminating the alignment signal. This also raises a deeper open question: correcting annotator bias may conflict with other alignment objectives, creating a fairness-alignment trade-off whose characterization remains an open problem.

Finally, these frameworks extend naturally to agentic AI systems \cite{xi2025rise}. Modern architectures increasingly rely on multi-agent pipelines where multiple components generate, evaluate, and aggregate information, a compositional structure that mirrors the ranking pipeline studied in this thesis. In such systems, bias may enter at each stage and compound across agents with heterogeneous distortions. When AI agents serve as evaluators, the bias parameters no longer represent human prejudice but the systematic distortions of AI systems, which may be subtler and more correlated than those of human annotators. Similarly, the abstention mechanisms introduced for ranking naturally generalize to agent delegation: determining when an agent should act versus defer to a human based on estimated risk. Developing fairness frameworks for these multi-agent dynamics, where unfairness emerges from the interaction of components rather than from any single one, represents a critical open challenge.

\clearpage

\pagestyle{plain}

\printbibliography

\end{document}